\documentclass{article}
\usepackage{arxiv}

\usepackage[utf8]{inputenc} %
\usepackage[T1]{fontenc}    %
\usepackage{hyperref}       %
\hypersetup{colorlinks,citecolor=blue,urlcolor=black,linkcolor=blue,pdfborder={0 0 0}}
\usepackage{url}            %
\usepackage{booktabs}       %
\usepackage{amsmath}        %
\usepackage{amsfonts}       %
\usepackage{nicefrac}       %
\usepackage{microtype}      %
\usepackage{algorithm}
\usepackage{algpseudocode}
\usepackage{graphicx}
\graphicspath{{media/}}     %
\usepackage{color}
\usepackage{xspace}
\usepackage{xcolor}
\usepackage{listings}
\usepackage{booktabs} %
\usepackage{multirow} %
\usepackage{tabularx}
\usepackage{authblk}
\RequirePackage[numbers,sort&compress]{natbib}

\usepackage{amsmath,amsfonts,bm}
\usepackage{thmtools}
\usepackage{mathrsfs}

\newcommand{\defeq}{\triangleq} %

\newcommand{\qtext}[1]{\quad\text{#1}\quad}

\def\eqref#1{equation~\ref{#1}}

\def\1{\bm{1}}

\def\rv{{\textnormal{v}}}

\DeclareMathAlphabet{\mathsfit}{\encodingdefault}{\sfdefault}{m}{sl}
\SetMathAlphabet{\mathsfit}{bold}{\encodingdefault}{\sfdefault}{bx}{n}

\newcommand{\R}{\mathbb{R}}

\newcommand{\approximate}[1]{\widehat{#1}}
\renewcommand{\rv}[1]{\uppercase{#1}}
\newcommand{\prob}[1]{\uppercase{#1}}

\newcommand{\fnfont}[1]{\mathsf{#1}}
\newcommand{\estfont}[1]{\mathtt{#1}}
\newcommand{\opfont}[1]{\mathscr{#1}}
\newcommand{\set}[1]{\mathcal{#1}}

\newcommand{\probset}[1]{\mathscr{P}(#1)}

\newcommand{\MMD}{\fnfont{MMD}}
\newcommand{\KSD}{\fnfont{KSD}}
\newcommand{\EP}{\fnfont{EP}}
\newcommand{\SIGReg}{\fnfont{SIGReg}}

\newcommand{\inputspace}{\set{X}}
\newcommand{\latentspace}{\set{Z}}
\newcommand{\loss}{\mathcal{L}}
\newcommand{\perturbfn}{\mathcal{T}}
\newcommand{\nperturbs}{m}
\newcommand{\regularizer}{\Omega}
\newcommand{\sg}{\fnfont{sg}}
\newcommand{\clstoken}{{\mathtt{[CLS]}}}

\newcommand{\mmd}[1][]{\MMD_{#1}}

\newcommand{\mmdgauss}[1][]{\estfont{MMD\text{-}Gauss}_{#1}}
\newcommand{\ummd}[1][]{\widehat{\fnfont{U}\text{-}\MMD}_{#1}}
\newcommand{\ksd}[1][]{\KSD_{#1}}

\newcommand{\ep}{\EP}

\newcommand{\kgsn}{k_{\mathrm{gsn}}}
\newcommand{\kimq}{k_{\mathrm{imq}}}

\newcommand{\imag}{\mathrm{i}}
\newcommand{\dif}{\mathrm{d}}

\newcommand{\sigreg}{\SIGReg}
\newcommand{\sigregest}{\estfont{SIGReg}}
\newcommand{\cwmmd}{\estfont{CWMMD}}
\newcommand{\cwksd}{\estfont{CWKSD}}

\newcommand{\sliceop}{\opfont{S}}
\newcommand{\direction}{\xi}

\newcommand{\twosampledir}[2]{\eta_{#2\to#1}}
\newcommand{\twosampledirest}[2]{\eta_{#1#2}}

\newcommand{\targetsym}{y}
\newcommand{\whitetarget}{\rv{\bar{\targetsym}}}

\newcommand{\basemetric}{\fnfont{d}}

\newcommand{\Normal}{\mathcal{N}}
\newcommand{\Unif}{\operatorname{Unif}}
\newcommand{\gaussshorthand}{\mathrm{N}}

\newcommand{\dd}{\mathop{}\!\mathrm{d}}
\newcommand{\iu}{\mathrm{i}}

\newcommand{\empirical}[1]{\widehat{#1}}
\newcommand{\empprob}[1]{\empirical{\rv{#1}}}
\newcommand{\learnprob}{\prob{p}}
\newcommand{\targetprob}{\prob{q}}
\newcommand{\sphere}[1][d]{\mathbb{S}^{#1 - 1}}

\newcommand{\kummer}{{}_1F_1}
\newcommand{\kummerkernel}[1][d]{\kappa_{#1}}

\newif\ifnotrestating
\notrestatingfalse

\newcommand{\prooflink}[1]{%
    {\small%
        \textup{\texttt{
            \ifnotrestating
                \hyperref[proof:#1]{[Proof]}
            \else
                \hyperref[claim:#1]{[Source]}
            \fi
        }}
    }
}

\makeatletter
\AtBeginEnvironment{restatable}{\notrestatingtrue}
\AtEndEnvironment{restatable}{\notrestatingfalse}
\makeatother

\newenvironment{talign*}
 {\let\displaystyle\textstyle\csname align*\endcsname}
 {\endalign}
\newenvironment{talign}
 {\let\displaystyle\textstyle\csname align\endcsname}
 {\endalign}

\usepackage{amsthm}
\usepackage{thmtools}
\usepackage{microtype}
\usepackage{nicefrac}
\usepackage{amssymb}
\usepackage[table]{xcolor} %
\usepackage[capitalize]{cleveref}
\crefname{equation}{}{}
\crefname{lemma}{Lem.}{Lems.}
\crefname{theorem}{Thm.}{Thms.}
\crefname{corollary}{Cor.}{Cors.}
\crefname{example}{Ex.}{Exs.}
\crefname{appendix}{App.}{Apps.}
\crefname{algorithm}{Alg.}{Algs.}
\crefname{algocf}{Alg.}{Algs.} %
\crefformat{section}{\S#2#1#3}
\crefrangeformat{section}{\S\S#3#1#4\ to\ #5#2#6}
\crefmultiformat{section}{\S\S#2#1#3}{ and\ #2#1#3}{, #2#1#3}{ and\ #2#1#3}
\crefname{table}{Tab.}{Tabs.}
\crefname{remark}{Rem.}{Rems.}
\crefname{definition}{Def.}{Defs.}
\crefname{proposition}{Prop.}{Props.}
\newtheorem{theorem}{Theorem}
\newtheorem{lemma}{Lemma}

\newtheorem{proposition}{Proposition}

\hypersetup{
    colorlinks=true,
    linkcolor=blue,
    citecolor=magenta,
}

\usepackage{multirow} %
\usepackage{tabularx}
\usepackage{booktabs} %
\usepackage{graphicx} %
\usepackage{listings}
\usepackage{xcolor}
\usepackage{float}
\usepackage{subcaption}

\definecolor{codegreen}{rgb}{0,0.6,0}
\definecolor{codegray}{rgb}{0.5,0.5,0.5}
\definecolor{codepurple}{rgb}{0.58,0,0.82}
\definecolor{backcolour}{rgb}{0.95,0.95,0.92}

\title{KerJEPA: Kernel Discrepancies for Euclidean\\Self-Supervised Learning}

\author[1]{Eric Zimmermann}
\author[2, 3]{Harley Wiltzer}
\author[2, 3]{Justin Szeto}
\author[1]{David Alvarez-Melis}
\author[1]{Lester Mackey}

\affil[1]{Microsoft Research, Cambridge, MA, United States}
\affil[2]{Mila--Qu\'ebec AI Institute, Montreal, Canada}
\affil[3]{McGill University, Montreal, Canada}

\begin{document}

\maketitle

\begin{abstract}
Breakthroughs in self-supervised Joint-Embedding Predictive Architectures (JEPAs) have established that regularizing Euclidean representations toward isotropic Gaussian priors yields provable gains in training stability and downstream generalization. We introduce a new, flexible family of Euclidean joint embedding self-supervised learning algorithms with kernel-based regularizers. One instance of this family corresponds to the SIGReg Epps-Pulley regularizer which approximates a sliced Maximum Mean Discrepancy (MMD) with a Gaussian prior and Gaussian kernel. By expanding the class of viable kernels and priors and computing the closed-form high-dimensional limit of sliced discrepancies, we develop alternative methods with improved training stability and design flexibility and leverage those techniques to train strong vision learners at scale. 
\end{abstract}

\section{Introduction}
Self-supervised learning (SSL) methods aim to learn useful, general representations from unlabeled data by learning the underlying structure of the data itself. The promise of these methods is that the representations serve as strong features for arbitrary downstream tasks, enabling the construction of foundation models. In this work, we study how the distribution imposed on Euclidean representations can be selected and estimated in a principled manner for visual data, allowing practitioners to make informed algorithm design decisions.

A common approach to SSL, particularly in visual domains, is to project data points, such as images, to various \emph{views}, such as random crops, and learn embeddings that map views of the same image to similar representations. A pitfall of this framework is that it admits a trivial, pathological solution that maps every input to the same embedding. To mitigate this issue, \emph{regularization} is crucial: by regularizing the distribution of the learned embeddings toward a fixed, user-designed \emph{prior} distribution, representation collapse can be avoided. However, regularizing such high-dimensional distributions remains a sincere challenge, both in principle and in practice.

Historically, SSL methods have largely constrained embeddings to a hypersphere \citep{simclr, align-unif}, a choice that stabilizes training and induces bounded gradients.
More recently, there has been increasing interest in moving off the hypersphere toward learning Euclidean representations \citep{vicreg,lejepa}.
In particular, LeJEPA \citep{lejepa} argues that an isotropic Gaussian distribution over embeddings is optimal for downstream regression and classification tasks with weight decay, a geometry that cannot be achieved by hyperspherical representations.
To induce this distribution, LeJEPA introduces SIGReg, a regularizer that compares randomly sampled univariate projections of the embedding distribution with univariate Gaussians.
This construction is principled, computationally attractive, and empirically effective.

Despite these successes, the reliance on stochastic projections introduces a fundamental tension between estimation variance and computational cost. The finite approximation of the expectation over directions introduces gradient noise, while the spectral integral within each projected discrepancy is itself approximated by numerical quadrature. Increasing the number of projections reduces the first source of error but eventually undermines the linear computational scaling that slicing is intended to provide in the standard SSL regime. Current sliced approaches specialize to Gaussian priors with a fixed Gaussian kernel, motivated by a result that isotropically-Gaussian-distributed features are optimal for downstream $\ell_2$-regularized tasks. Unfortunately, this ignores the distribution shift between the regularized features and those used for transfer in SSL pipelines.
This rigidity limits the exploration of alternative regularizers, non-Gaussian priors, and heavy-tailed kernels which may offer superior geometric properties for downstream tasks. 

In this work, we establish an equivalence between SIGReg and a specific kernel discrepancy; generalizing the kernel and the discrepancy reveals a broader family of distributional regularizers for self-supervised learning. This perspective enables normalization-free representation learning, while providing a principled framework to study alternative discrepancy constructions and their computational and statistical tradeoffs. In particular, we investigate finite-direction slicing and analytically integrated estimators, and apply such methods to larger scale datasets. Our contributions are summarized as follows:
\begin{itemize}
    \item \textbf{Theoretical foundation.} We show that SIGReg is a special case of a broader family of kernel discrepancies, providing a unified perspective on distributional regularization in SSL.
    \item \textbf{Regularizer design space.}
    We derive algorithms induced by alternative estimators for this family of regularizers, comparing statistical and computational tradeoffs at scale.
    \item \textbf{Scalable training.}
    We achieve strong ImageNet performance with scalable student-teacher vision transformers, outperforming comparable models trained on identical data scales.
\end{itemize}
\section{Related Works}
\label{sec:related}

While the broader goal of SSL is to learn augmentation-invariant representations, methods diverge in how they prevent representation collapse. \emph{Contrastive} methods, such as SimCLR \citep{simclr} and MoCo \citep{moco}, avoid collapse by explicitly pushing apart the representations of distinct samples (negative pairs) while attracting positive pairs. \emph{Non-contrastive} methods eliminate the need for negative pairs, relying instead on architectural asymmetries or statistical constraints. For instance, BYOL \citep{byol} utilizes asymmetric stop-gradient mechanisms, while Barlow Twins \citep{barlow} and VICReg \citep{vicreg} impose variance-preservation and de-correlation objectives on the embedding dimensions.

Crucial to all these frameworks is the design of data augmentations, which implicitly defines which features are learned. Techniques such as cropping, color jittering, and Gaussian blur encourage the network to discard low-level visual statistics \citep{domain-ssl}, while advanced strategies like multicrop \citep{swav} enforce consistency across varying spatial scales. Equally important is the standard backbone-projector architectures. Empirical success in SSL heavily relies on guillotine regularization, where the non-linear projection head is discarded after pretraining, while only the backbone features are used for downstream tasks \citep{guillotine}. This structural decoupling implies that the training objective, which is applied to the projections, does not directly constrain the geometry of the representations used for transfer. This misalignment draws attention to the benefits of selecting a particular geometry throughout training.

A dominant paradigm in SSL involves normalizing embeddings to the unit hypersphere, effectively fixing the scale of representation to focus learning entirely on angular relationships. This geometric constraint naturally leads to objectives that encourage the uniform distribution of features over the sphere, thereby maximizing the information entropy of the induced distribution \citep{align-unif}. More sophisticated distribution matching has been approached through optimal transport, specifically using the Spherical Sliced Wasserstein (SSW) distance \citep{spheresliced}, which leverages random projections to align the empirical distribution of embeddings with a uniform target. Similarly, kernel-based methods have been applied via Hilbert-Schmidt Independence Criterion \citep{hsic-ssl}. These works demonstrated the utility of both slicing and kernel discrepancies, providing a foundation for their extension into other geometries.

The recent shift toward isotropic Gaussian targets in Euclidean space, exemplified by LeJEPA \citep{lejepa} which builds on VICReg \citep{vicreg, vicreg-mmd}, bridges SSL with kernel-based methods, without explicit normalization. Although LeJEPA formulates its regularizer via the \emph{Epps-Pulley test} \citep{ep-test} on random projections, recent analyses establish that this is equivalent to minimizing the Sliced Maximum Mean Discrepancy between the embeddings and a target Gaussian \citep{metrics-prob-measures, kernel-gof}.
Based on this categorization, there is a plethora of novel discrepancies and estimators waiting to be explored \citep{ksd, ksdd, sksd}, which we investigate in the sequel.

\section{Background}
\label{sec:background}
This paper studies a broad family of self-supervised learning algorithms, by carefully treating the problem of regularizing learned representations via distributional similarity estimation. In SSL, the goal is to learn an embedding function $f_\theta : \inputspace \to \latentspace$ which maps inputs from space $\inputspace$ (say, high-dimensional images or videos) to a (lower dimensional) latent space $\latentspace\subseteq\R^d$, such that the representations enable efficient downstream processing (for instance, accurate linear classification).

Leveraging known invariance structure of the data, such a feat can readily be accomplished. Suppose $\perturbfn$ is an augmentation map, which maps any input $x$ to a distribution over \emph{views} of $x$, which could be random crops or rotations that representations should be invariant to.
Then SSL can be accomplished by minimizing a squared loss between predicted representations perturbed from the same source input, subject to regularization of the population of predicted representations \emph{across} source inputs relative to a fixed target.
This is encapsulated by the following loss $\loss$, where $\{\rv{x}_l\}_{l=1}^n \subseteq \inputspace$ is a minibatch of inputs, and $\nperturbs$ is the number of perturbations applied to each source input, 
\begin{talign}\label{eq:ssl:loss:generic}
    \begin{aligned}
        \!\!\!\!\loss(\theta)
        &\defeq \frac{1}{dn\nperturbs^2}\sum_{l=1}^n\sum_{i,j=1}^\nperturbs \Vert f_\theta (\rv{v}^i_l) - f_\theta (\rv{v}^j_l)\Vert^2_2 + \lambda \regularizer(\empprob{p}, \prob{q}),\ 
        \empprob{p} \defeq \frac{1}{n\nperturbs}\sum_{l=1}^n\sum_{i=1}^\nperturbs \delta_{f_\theta (\rv{v}^i_l)}\hspace{-1em}
    \end{aligned}
\end{talign}
where each $\rv{v}_l^i$ is sampled independently from $\perturbfn(\rv{x}_l)$.
In \cref{eq:ssl:loss:generic}, $\prob{Q}$ is a fixed \emph{prior} over latent variables (say, a uniform distribution in the case of bounded latent space, or a Gaussian distribution), $\empprob{Z}$ is an empirical distribution of latents output by the model $f_\theta$, and $\regularizer$ is a regularization between latent distributions minimized when the inputs are equal.
Notably, this regularization term is crucial: without it, \cref{eq:ssl:loss:generic} could be minimized by trivial solutions such as constant $f_\theta$.

Designing a powerful and efficient regularizer remains an active direction of research in SSL.
Traditionally, SSL methods have constrained latent representations to the sphere $\latentspace = \sphere$, which benefits from enforcing bounded representations useful for downstream application, but suffers from complications during training.
In response to these complications, \citet{lejepa} introduced the \emph{LeJEPA} instantiation of \cref{eq:ssl:loss:generic} for Euclidean representations $\latentspace = \R^d$.
This instantiation notably introduces the regularizer $\Omega = \sigreg$ and Gaussian prior $\prob{Q}=\Normal(0, \sigma^2)$, citing various theoretical justifications and practical simplicity and effectiveness.

The $\sigreg$ regularizer breaks the curse of dimensionality involved in estimating distributional distance in $\probset{\R^d}$ by computing $\Theta(1)$ random scalar projections of $\empprob{Z}$ and computing the \emph{Epps-Pulley} test statistic via quadrature of an integral on the real line; a principled method with provable statistical guarantees.
In this paper, we argue that \textbf{$\sigreg$ is not as fundamental as it may seem}. Rather, it is built on various building blocks (namely, projection, Epps-Pulley, and quadrature) which are merely a particular means of computing the special case of a much more general class of regularizers.
The remainder of the paper will delve into this more general class, providing efficient algorithms that generalize LeJEPA, and offering computational and practical benefits.
The remainder of this section discusses kernel-based distance analysis methods, and the building block of LeJEPA.

\subsection{Maximum Mean Discrepancy and Kernel Stein Discrepancy}
\label{sec:mmd}
\label{sec:ksd}
At the core of our investigation in this paper is the \emph{Maximum Mean Discrepancy} \cite[MMD]{kernel-two-sample}, a kernel-based distance between distributions.
Given a kernel $k:\latentspace\times\latentspace\to\R$ measuring similarity between \emph{samples}, the MMD between distributions $\prob{p}, \prob{q}$ is given by
\begin{talign}
\mmd[k]^2(\prob{P}, \prob{Q})
    =
\mathbb{E}_{\rv{x},\rv{x}' \sim \prob{P}}[k(\rv{x},\rv{x}')] 
+ \mathbb{E}_{\rv{y},\rv{y}' \sim \prob{Q}}[k(\rv{y},\rv{y}')] 
- 2\mathbb{E}_{\rv{x} \sim \prob{P}, \rv{y} \sim \prob{Q}}[k(\rv{x},\rv{y})].
\label{eqn:mmd-expanded}
\end{talign}
The choice of kernel determines a notion of similarity between embeddings.
For particular \emph{characteristic} kernels, such as the Gaussian kernel, $\MMD_k(P,Q)$ vanishes exactly when $\prob{p}=\prob{q}$ \citep{kernel-two-sample,metrics-prob-measures}.
Common examples of characteristic kernels include Gaussian kernels $\kgsn$ and inverse multi-quadric kernels $\kimq^{\alpha,\beta}$, given by
\begin{talign*}
    \kgsn(x, y) \defeq \exp(-\gamma\|x-y\|_2^2)
    \quad\text{and}\quad
    \kimq^{\alpha,\beta}(x, y) \defeq (1 + \alpha\|x - y\|_2^2)^{-\beta}
\end{talign*}
with $\gamma,\alpha,\beta>0$.
Notably, \cref{eqn:mmd-expanded} admits an unbiased estimator by drawing samples from both distributions and averaging kernel evaluations between sample pairs.

The \emph{Kernel Stein Discrepancy} (KSD) is an alternative kernel-based distance between distributions,  defined through the target score \(s_{\prob{Q}}(x)=\nabla_x\log q(x)\) \citep{kernel-gof,ksd,sample-quality-kernels}.
For a base kernel $k$, comparable to the kernel discussed for the MMD, the KSD constructs a kernel $k_Q$ through the \emph{Stein operator}, and is given by
\begin{talign}
\small
\begin{aligned}
\ksd[k]^2(\prob{P},\prob{Q})
&\defeq
\mathbb{E}_{\rv{x},\rv{x}'\sim\prob{P}}
\left[
k_{\prob{Q}}(\rv{x},\rv{x}')
\right],\quad \text{where}\\
k_{\prob{Q}}(x, y) &\defeq s_{\prob{Q}}(x)^\top k(x, y) s_{\prob{Q}}(y) + s_{\prob{q}}(x)^\top \nabla_y k(x, y) + \nabla_x k(x, y)^\top s_{\prob{q}}(y) + \mathrm{tr}(\nabla_x\nabla_y^\top k(x, y)).\hspace{-0.5em} 
\end{aligned}
\label{eqn:ksd-expanded}
\end{talign}
By Stein's identity, KSD is equivalent to MMD with a target-dependent kernel, where the expectation is taken only under \(\prob{P}\) and evaluations of the target score  \citep{ksd,barp2024targeted}.

\subsection{Spectral Representations of Discrepancies}
\label{sec:spectral}
Shift-invariant kernels admit equivalent frequency-domain representations via Bochner's Theorem \citep{bochner}. Let \(\rho_k(\omega)\) denote the spectral density associated with \(k\) and let $\phi_Q(\omega)$ denote the characteristic function (CF) of a known distribution $\prob{Q}$. Given samples \(\rv{x}_1,\ldots,\rv{x}_n\overset{\mathrm{iid}}{\sim}\prob{P}\), and the empirical distribution \(\empprob{P}_n=n^{-1}\sum_{i=1}^n\delta_{\rv{x}_i}\), MMD and KSD are written using the empirical characteristic function (ECF) defined as $\widehat{\phi}_n(\omega)= n^{-1}\sum_{i=1}^{n}
\exp(\iu\omega^\top\rv{x}_i)$ and its score-weighted counterpart written as $\widehat{\phi}_{s_{\prob{Q}},n}(\omega) =n^{-1}\sum_{i=1}^{n}
s_{\prob{Q}}(\rv{x}_i)\exp(\iu\omega^\top\rv{x}_i)$. Using these two empirical spectral features, MMD and KSD take the parallel forms
\begin{talign*}
    \begin{aligned}
        \mmd[k]^2(\empprob{P}_n,\prob{Q})
        &\defeq
        \int_{\mathbb{R}^d}
        |
        \widehat{\phi}_n(\omega)-\phi_{\prob{Q}}(\omega)
        |^2
        \rho_k(\omega)\,\dd\omega, \\
        \ksd[k]^2(\empprob{P}_n,\prob{Q})
        &\defeq
        \int_{\mathbb{R}^d}
        \|
        \widehat{\phi}_{s_{\prob{Q}},n}(\omega)
        +\iu\omega\widehat{\phi}_n(\omega)
        \|^2
        \rho_k(\omega)\,\dd\omega.
    \end{aligned}
\end{talign*}
These spectral representations are useful because the frequency integrals can be approximated with deterministic quadrature or Monte Carlo frequencies, including Random Fourier Features, without pairwise kernel evaluations \citep{rff,mmd-rff,random-stein-feats}.

\subsection{Sliced Discrepancies}
\label{sec:sliced}
Certain distances between multi-dimensional distributions are computationally or statistically difficult to estimate, succumbing to the curse of dimensionality.
The spectral representations of discrepancies in \cref{sec:spectral}, for instance, require integrating over $\R^d$, which is exponentially expensive to estimate.

The concept of \emph{slicing} or random projection can alleviate several computational challenges arising from the sample-based estimation of multivariate random variable discrepancies.
Explicitly, such approaches leverage the celebrated \emph{Cram\'er-Wold} theorem, which asserts that $\direction^\top\rv{x}$ has the same distribution as $\direction^\top\rv{y}$ for any direction $\direction\in\mathbb{S}^{d-1}$ whenever $\rv{x}$ has the same distribution as $\rv{y}$.
This motivates the notion of \emph{sliced divergences}, which lift divergences on univariate distributions to multivariate distributions through the \emph{slice operator} $\sliceop$, given by $(\sliceop\basemetric)(\prob{p}, \prob{q}) \defeq \mathbb{E}_{\direction\sim\Unif(\mathbb{S}^{d-1})}[\basemetric(\direction^\top_\#\prob{p}, \direction^\top_\#\prob{q})]$.

As an immediate note, whenever a metric $\basemetric$ admits an unbiased sample estimator, $\sliceop\basemetric$ does as well: simply sample $\nperturbs$ directions $\direction$ independently and uniformly on $\mathbb{S}^{d-1}$, and then average the unbiased estimates $\approximate{\basemetric}(\direction^\top_\# \learnprob)$ of $\basemetric(\direction^\top_\#P, \direction^\top_\#Q)$. For a base divergence $\basemetric$ with sample complexity $\beta(d, n)$ for $n$ samples drawn from the probability distributions under comparison, this Monte-Carlo sliced distance estimator has sample complexity bounded \cite[Corollary S4]{nadjahi2020statistical} by
\begin{talign*}
    \beta(d, n) + \sqrt{\frac{\mathbb{E}\left[\mathrm{Var}\left\{\approximate{\basemetric}(\rv{\direction}^\top_\# \learnprob)\right\}\right]}{m}}.
\end{talign*}

\subsection{The Epps-Pulley Test and LeJEPA}
\label{sec:lejepa}
Having defined several forms of distribution distances, the $\sigreg$ regularizer from LeJEPA \citep{lejepa} can be formulated.
$\sigreg$ is built upon a quadrature estimate of the \emph{Epps-Pulley} method \citep{ep-test}, given by 
\begin{talign*}
    \EP(\prob{p}; \Normal(0, \sigma^2)) \defeq \int_{\R}|\phi_{\prob{P}} - \phi_{\Normal}|^2 \rho(\omega)\dd\omega,
\end{talign*}
where $\phi_{\Normal}$ is the ECF of $\Normal(0, \sigma^2)$ for some $\sigma > 0$, and $\rho$ is the spectral density of a Gaussian kernel with bandwidth $\gamma > 0$.
Notably, $\EP$ is an instance of the MMD with a univariate Gaussian kernel.
Consequently, multidimensional extensions can be made by employing MMD estimators with a multivariate Gaussian kernel as well; a notable example is the BHEP test \citep{Baringhaus1988}.
In effect, $\sigreg$ is the $\EP$ specialized to the target $\prob{q} = \Normal(0, \sigma^2 I_d)$, given by
\begin{talign*}
    \regularizer(\prob{p}, \Normal(0, \sigma^2)) \defeq \sigreg(\prob{p}) = (\sliceop \EP)(\prob{p}; \Normal(0, \sigma^2 I_d)).
\end{talign*}
In practice, LeJEPA estimates $\sigreg$ via Monte Carlo over the slicing operation as discussed in \cref{sec:sliced}, and with quadrature on the 1-dimensional $\EP$ computations.

\textbf{Notationally}, it will be convenient to easily distinguish regularizers from their estimators; as such, regularizers will be written in $\fnfont{sans\text{-}serif}$, while estimators are written in $\estfont{monospace}$.
As such, $\sigregest$ will refer to the instantiation of the \emph{estimator} for $\sigreg$ in LeJEPA, constructed by performing quadrature estimation with a finite set of anchor points.

\section{General Euclidean Regularization by Distribution Matching}
\label{sec:euclidean}
As shown \cref{sec:lejepa}, the SIGReg regularizer in LeJEPA defines a metric between feature distributions for preventing feature collapse, but also prescribes a \emph{method} for estimating this metric, notably with slicing and quadrature estimation.
Recognizing this distinction removes layers of complexity due to the estimation method imposed by LeJEPA, unveiling a simple and familiar regularizer at the core.

\begin{restatable}{theorem}{kummermmd}\label{claim:kummermmd}
    Let $\sigreg$ denote the SIGReg regularizer \citep{lejepa}; that is
    \begin{talign*}
        \sigreg(\learnprob) \defeq  \mathbb{E}_{\rv{\direction}\sim\Unif(\mathbb{S}^{d-1})}[n^{-1}\ep(\rv{\direction}^\top_\#\learnprob)],\quad\prob{p}\in\probset{\R^d}.
    \end{talign*}
    It holds that $\sigreg$ is equivalent to a particular MMD,
    \begin{talign*}
        \sigreg(\learnprob) = \mmd[\kummerkernel]^2(\learnprob, \mathcal{N}(0,\sigma^2I_d))
        \qtext{where} \kummerkernel(x, y) \defeq  \kummer\left(\frac{1}{2}; \frac{d}{2}; -\gamma\|x - y\|_2^2\right),
    \end{talign*}
    and $\kummer$ is the \emph{Kummer Confluent Hypergeometric function}.
    In particular, $\kummerkernel$ has polynomially decaying tails (as opposed to exponentially decaying tails like $\kgsn$).
    \hfill\prooflink{kummermmd}
\end{restatable}
\cref{claim:kummermmd} sheds light on the induced \emph{geometry} imposed on the latent embeddings by the slicing operation. Notably, it imposes a \emph{heavy tailed} similarity measure between latent embeddings, as opposed to commonly used subgaussian-tailed kernels such as $\kgsn$.
Additionally, the equivalent kernel induced by slicing is \emph{dimension-dependent}: as we scale the dimension, the slicing operation effectively decreases the \emph{bandwidth} of the resulting kernel. As a result, it is important to consider both the choice of kernel and the use of slicing when designing distributional regularizers in high dimensional settings. The geometry induced by slicing motivates direct use of the IMQ kernel as an explicit heavy-tailed alternative in the ambient space without any additional slicing, while its dimension-dependent bandwidth suggests a simple rescaling may be sufficient. We therefore examine both sliced and non-sliced discrepancies with Gaussian and IMQ kernels experimentally, allowing us to assess the true importance of kernel tail behavior in the context of a self-supervised regularizer.

Next, we turn our attention back to estimation of the regularizer.
Since \cref{claim:kummermmd} effectively reduces $\sigreg$ to a particular MMD, we may turn to standard MMD estimation techniques \citep{kernel-two-sample} to derive alternative estimators with different computational and statistical tradeoffs.
\begin{restatable}{theorem}{kummermmdsample}
\label{claim:kummermmdsample}
Let $\learnprob\in\probset{\mathbb{R}^d}$, and let $\empprob{P}_n \defeq  (\rv{x}_i)_{i=1}^n\overset{\mathrm{iid}}{\sim} \learnprob$.
Define the estimator $\mmdgauss[\kummerkernel]$ via
\begin{talign*}%
    \mmdgauss[\kummerkernel](\empprob{P}_n)
    \defeq  \frac{1}{n(n-1)}\sum_{i=1}^n\sum_{j\neq i}\kummerkernel(\rv{x}_i, \rv{x}_j)
    - \frac{2}{n\sqrt{1 + 2\gamma\sigma^2}}\sum_{i=1}^n\kummer(\frac{1}{2};\frac{d}{2};-\frac{\gamma}{1 + 2\gamma\sigma^2}\|\rv{x}_i\|^2_2) + C
\end{talign*}
for some constant $C$ independent of $\learnprob$.
Then $\mmdgauss[\kummerkernel](\empprob{P}_n)$ is an unbiased estimate of $\sigreg(\learnprob)$, and $\lvert \mathbb{E}[\mmdgauss[\kummerkernel](\empprob{P}_n) - \sigregest(\empprob{P}_n)]\rvert \leq O(1/n)$.
\hfill\prooflink{kummermmdsample}
\end{restatable}
The estimator of \cref{claim:kummermmdsample} presents interesting tradeoffs relative to the sliced Epps-Pulley estimator of LeJEPA \citep{lejepa}.
Both estimators require samples
$\empprob{P}_n$ from $\learnprob$. The tradeoffs lie in which additional random samples are required, as well as computational efficiency.
Notably, the $\mmdgauss[\kummerkernel]$ estimator avoids Monte-Carlo sampling slice directions by effectively computing the expectation over directions in closed form: this is what brings us from the simple $\kgsn$ kernel to $\kummerkernel$ and $\kummer$ evaluations.
Moreover, in principle, $\mmdgauss[\kummerkernel]^2$ is \emph{unbiased}, unlike the implementation of LeJEPA which requires approximating a Fourier integral via quadrature methods.
However, these advantages come at a computational cost, which we discuss next.

In transitioning from sampling directions and evaluating univariate MMDs (e.g., Epps-Pulley tests), we perform pairwise kernel evaluations, costing $O(n^2)$ computation as opposed to the $O(n)$ complexity of LeJEPA.
That said, it is worth emphasizing that this is but a design decision as opposed to an algorithmic necessity.
Indeed, there is a rich literature on MMD estimation, and linear-time unbiased estimators are well-established \citep[Lemma 14]{kernel-two-sample}, though these generally come with higher variance.
We leave the exploration of such estimators for future work.

\subsection{A Slice-Free Tractable Estimator}\label{sec:euclidean:mmd}
In this section, we gave an unbiased estimator that estimates the same regularization as LeJEPA's SIGReg, without necessitating random slices.
Unfortunately, in doing so, an additional complication snuck in, since $\kummer$ and thus the slice-induced kernel $\kummerkernel$, are generally intractable to compute exactly.
Fortunately, due to the following result of \citet{cwae}, this kernel can be well approximated by a simpler characteristic kernel in the high-dimensional regime.
\begin{theorem}[\citet{cwae}, Proposition 4]\label{thm:kummer-to-imq}
It holds that
\begin{talign*}
    \kummer\left(\frac{1}{2}; \frac{d}{2}; -c\right) - \kimq^{\nicefrac{4}{2d-3},\nicefrac{1}{2}}(c)
    \overset{d\uparrow\infty}{\longrightarrow} 0
    \qtext{for}
    \kimq^{\nicefrac{4}{2d-3},\nicefrac{1}{2}}(c)
    \defeq
    \frac{1}{\sqrt{1 + \frac{4}{2d-3}c}}.
\end{talign*}
As a consequence, it follows immediately that $\kummerkernel = \kimq^{\nicefrac{4\gamma}{2d-3},\nicefrac{1}{2}} + o(1)$.
\end{theorem}
While \citet{cwae} doesn't provide a quantitative convergence rate, they note that the approximation tends to be practically very useful once $d \geq 20$.
This is especially relevant in our setting, where the goal is to train large generalist models, where the feature dimension is often greater than $20$ by at least an order of magnitude.
This motivates the following tractable $\cwmmd$ estimator for $\sigreg$.
\begin{restatable}{theorem}{thmcwmmdest}\label{claim:thmcwmmdest}
    Let $\prob{p}\in\probset{\R^d}$, and let $\empprob{p}_n \defeq  (\rv{x}_i)_{i=1}^n \overset{\mathrm{iid}}{\sim} \prob{p}$. Define the estimator $\cwmmd$ via
    \begin{talign*}
        \cwmmd(\empprob{p}_n) \defeq  \frac{1}{n(n-1)}\sum_{i=1}^n\sum_{j\neq i}\kimq^{\nicefrac{4\gamma}{2d-3},\nicefrac{1}{2}}(\rv{x}_i, \rv{x}_j) - \frac{2}{n\sqrt{1 + 2\gamma \sigma^2}}\sum_{i=1}^n\frac{1}{\sqrt{1 + \frac{4\gamma}{(2d-3)(1 + 2\gamma\sigma^2)}\|\rv{x}_i\|_2^2}} + C.
    \end{talign*}
    Then $\lvert \mathbb{E}[\cwmmd(\empprob{P}_n) - \sigregest(\empprob{P}_n)]\rvert \leq O_d(1/n)$.
    \hfill\prooflink{thmcwmmdest}
\end{restatable}
Altogether, both $\cwmmd$ and LeJEPA induce a small bias in regularization, and the former expends quadratic compute to eliminate the variance of Monte-Carlo slicing.
With large distributed training setups, we argue that it is worth investigating which side of this tradeoff is favorable, or identifying situations when one approach may be preferred over the other.
That said, it is important to note that \textbf{slicing divergences is not necessary or sufficient for breaking the curse of dimensionality}. We showed that equivalent regularizers can be derived (and indeed, generalized) without slicing at all.

\subsection{A Novel Kernel Stein Regularizer}\label{sec:euclidean:ksd}
So far we have separated the regularization metric from the estimator introduced by LeJEPA, revealing alternative slice-free estimators leveraging kernel discrepancies with heavy-tailed kernels.
Particularly, we showed that $\sigreg$ is equivalent to MMD regularization with a particular kernel.
In this section, we apply a similar technique to \cref{sec:euclidean:mmd} to derive an equivalent estimator for the sliced KSD \citep{sksd}, which can be computed without random projections.
\begin{restatable}{theorem}{thmcwksdest}\label{claim:thmcwksdest}
    Let $\prob{p}\in\probset{\R^d}$, and let $\empprob{p}_n \defeq  (\rv{x}_i)_{i=1}^n \overset{\mathrm{iid}}{\sim} \prob{p}$. Define $\twosampledirest{i}{j} \defeq  \|\rv{x}_i - \rv{x}_j\|^{-1}(\rv{x}_i - \rv{x}_j)$, and consider the estimator $\cwksd: \R^{d\times n}\to\R_+$ given by
    \begin{talign*}
        \cwksd(\empprob{p}_n)
        &\defeq  \frac{1}{n(n-1)}\sum_{i=1}^{n}\sum_{j\neq i}\Big(2\gamma + \frac{\rv{x}_i^\top \rv{x}_j - \twosampledirest{i}{j}^\top \rv{x}_i\twosampledirest{i}{j}^\top \rv{x}_j}{\sigma^4(d-1)}\Big)\kimq^{\nicefrac{4\gamma}{2d-3},\nicefrac{1}{2}}(\rv{x}_i, \rv{x}_j)\\
        &\qquad+ \frac{1}{n(n-1)}\sum_{i=1}^n\sum_{j\neq i}\frac{1}{d}\Big(\frac{\twosampledirest{i}{j}^\top \rv{x}_i\twosampledirest{i}{j}^\top \rv{x}_j}{\sigma^4} - \frac{\rv{x}_i^\top \rv{x}_j - \twosampledirest{i}{j}^\top \rv{x}_i\twosampledirest{i}{j}^\top \rv{x}_j}{\sigma^4(d-1)}\Big)\kimq^{\nicefrac{2\gamma}{d}, \nicefrac{3}{2}}(\rv{x}_i, \rv{x}_j)\\
        &\qquad+ \frac{1}{n(n-1)}\sum_{i=1}^n\sum_{j\neq i}\frac{1}{d}\big(\frac{2\gamma}{\sigma^2} + 4\gamma^2\big)\|\rv{x}_i - \rv{x}_j\|_2^2\kimq^{\nicefrac{2\gamma}{d}, \nicefrac{3}{2}}(\rv{x}_i, \rv{x}_j).
    \end{talign*}
    Then $\mathbb{E}[\cwksd(\empprob{p}_n)] = \sliceop\ksd[\kgsn]^2(\prob{p}, \Normal(0, \sigma^2I_d)) + o_d(1)$.
    \hfill\prooflink{thmcwksdest}
\end{restatable}
\cref{claim:thmcwksdest} shows that the sliced $\KSD^2$ can be estimated without sampling random projections, by instead averaging heavy-tailed kernel evaluations on latent samples which are already present. Like $\cwmmd$, variance of random projections is traded off for quadratic compute in the number of samples $n$.
\section{Methods}
\label{sec:methods}

\textbf{Network architecture}.\label{sec:network}
KerJEPA is composed of a Vision Transformer (ViT) \citep{Kolesnikov2021} and three layer projector of hidden dimension of 4096 and GELU activations with output dimension of 256. For the ViT-S, we divide these values in half. During inference, the projector is discarded. Use of $\clstoken$ or patch tokens are task dependent. 

\textbf{Target standardization}.\label{sec:target-standardization}
To preserve the statistical power of the chosen kernel over the course of training, we introduce a target standardization procedure inspired by \citet{pmlr-v139-ermolov21a}. This substitutes pairwise MSE in \cref{eq:ssl:loss:generic} with MSE against an aggregated target,
\begin{talign*}
    \!\!\!\loss(\theta) \defeq \frac{1}{dn\nperturbs}\sum_{l=1}^n \sum_{i=1}^\nperturbs \Vert f_\theta(\rv{v}_l^i) -  \sg(\whitetarget_l)\Vert^2_2 + \lambda \regularizer(\empprob{p}, \prob{q}),
    \ \ 
    \whitetarget_l \defeq \frac{1}{\empirical{\sigma}}{(\frac{1}{\nperturbs}\sum_{i=1}^\nperturbs f_\theta(\rv{v}^j_l) - \empirical{\mu}}){},
\end{talign*}
where $\sg$ is the stop-gradient operator, and where $\empirical{\mu}, \empirical{\sigma}$ are derived from EMA estimates of the global mean and variance of features \citep{caron2021emerging}.
By incorporating standardization directly into the alignment target, we constrain the location and scale of the representation space, discouraging trivial collapse while reducing sensitivity to the kernel bandwidth.

\textbf{Implementation details}.\label{sec:implementation}
We pretrain KerJEPA without labels in two settings. First, we train a ViT-S/8 on Imagenette \citep{imagenette} for 800 epochs using AdamW \citep{loshchilov2019decoupled}, a cosine learning rate schedule with a peak lr of $5\times10^{-4}$, weight decay of $5\times10^{-2}$, and a batch size of 256. Each sample consists of 4 global views of size $128\mathrm{px}^2$. We evaluate both MMD and KSD across a range of kernels and priors in the sliced, and unsliced setting, and analyze stability and convergence properties. We provide pseudocode for sliced regularizers in \cref{app:kerjepa-pseudo}.

Second, we evaluate three model scales with patch size 16 on ImageNet using a target normalized self-distilled student-teacher framework with EMA, scheduled cosine-wise from $0.992$ to $1.0$. We fix the loss to $\cwmmd$ with a weight of $\lambda =7.5$, or $\cwksd$ with a weight of $\lambda=2.5$ and select the bandwidth using the one-dimensional Silverman's Rule \citep{Silverman2018}. We use 2 global views of size $224\mathrm{px}^2$, 10 local views of size $96\mathrm{px}^2$, and a batch size of 1024. Weight decay is warmed up from $0.05$ to $0.6$ for ViT-S/B and $0.8$ for ViT-L over the first $75\%$ of training, while learning rate decays from $8\times10^{-4}$ to $1.5\times10^{-4}$. We apply a $0.2$ learning rate scaling to the patch-convolution and layerwise decay of 0.9 for stability \citep{chen2021mocov3,oquab2024dinov}. In the multi-crop setting, the target is constructed exclusively from global views, and representations from both local and global views are aligned to this standardized target. ViT-S/B/L models are trained for 800, 400, and 250 epochs .
Across settings, learning rate is warmed up on the first 10 epochs, and we use augmentations following BYOL \citep{byol} and SWAV \citep{swav}. Regularization is only applied to global views and are averaged.

\textbf{Evaluation protocols}.\label{sec:eval-protocol}
We evaluate frozen backbone features using linear probing and $k$-nearest-neighbor ($k$-NN) classification. For linear probing, a classifier is trained online with random resized crops and horizontal flips, while evaluation uses center crops.
We sweep learning rates $\{10^{-k}, 5\times 10^{-k}\}_{k=2}^5$ and use no weight decay.
For $k$-NN, classification is performed offline on unaugmented features, sweeping $k \in \{10, 20, 50, 100, 200\}$. We restrict comparisons to strong open-weight ImageNet-trained models from DINO \cite{caron2021emerging} and iBOT \citep{zhou2022ibot}, standardizing architectures and training data.
\section{Experimental Results}
\label{sec:main-results}

We organize our experiments around the progression from the theoretical properties of our discrepancy estimators to their practical performance and design choices. We first validate the predicted dependence of finite-slice estimation on representation dimensionality, showing that insufficient slicing leads to slower and less stable convergence. We then demonstrate the practical benefit of the resulting analytic formulations on ImageNet-1k, where KerJEPA learns strong visual representations across model scales. Finally, we examine the design and optimization choices that contribute to performance and stable training. We call unsliced methods those which do not apply projections and compute the discrepancy in the ambient space, sliced methods as those which compute the Monte-Carlo estimation of the discrepancy under finite slices, and the analytically sliced as those which compute the discrepancy as the limiting case of all slices.

\begin{figure}[h]
    \centering
    \includegraphics[trim=0 .1cm 0 0,clip,width=1\linewidth]{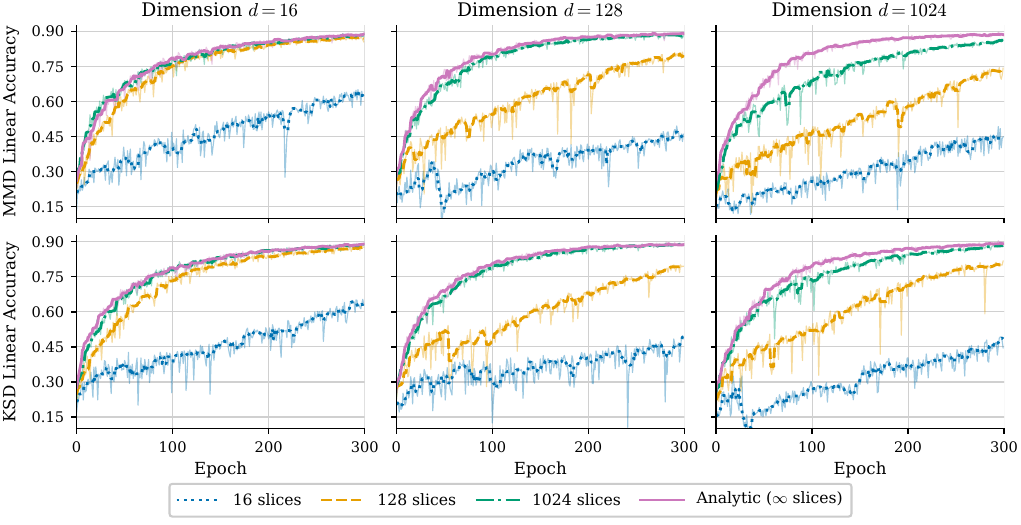}
    \vspace{-\baselineskip}
    \caption{Impacts of slicing in various dimensions for the MMD and KSD regularizer on ImageNette, measuring test accuracy over various output dimensions. Insufficient slices slow convergence rates as a function of dimension and introduces instability in the early half of training.}
    \label{fig:slice-scaling}
\end{figure}

\textbf{Slicing requirements.} Finite-slice estimators introduce Monte Carlo noise whose effect depends on the dimensionality of the output space. As the number of dimensions increases, more slices are required to obtain a sufficiently accurate approximation. \cref{fig:slice-scaling} confirms this behavior and demonstrates how insufficient numbers of slices increasingly slow convergence and introduce instability during the early stages of training. In contrast, the closed-form estimators for $\cwmmd$ and $\cwksd$ avoid this source of Monte Carlo error and exhibit faster and more stable convergence across dimensions. These results motivate the use of the analytic formulations for our high-dimensional experiments.

\textbf{Discrepancy design spaces.} We examine the impacts of different discrepancies, kernels, and priors across sliced, unsliced, and analytically sliced options. \cref{tab:ablation-results} shows that all priors and approximation methods converge to similar results, despite their training trajectories. The framework is thus not tied to a single kernel, prior, or slicing requirement. See \cref{app:kerjepa-pseudo} for examples of practical sliced algorithms.
\begin{table}[h]
    \centering
    \caption{\textbf{ImageNette regularization ablations.} Top-1 accuracy (\%) across different settings on models trained for 800 epochs with dimension of 128. Unsliced heavy-tail kernels perform best.}
    \begin{tabular}{l l l l c c c}
        \toprule
        \textbf{Discrepancy} & \textbf{Type} & \textbf{Base Kernel} & \textbf{Prior} & \textbf{Slices} & \textbf{Linear}\\
        \midrule
        \midrule
        \multirow{3}{*}{\textbf{MMD}} 
            & Unsliced & Gaussian & Gaussian & -- & 91.29 \\
            & Sliced (Analytic) & Gaussian & Gaussian & $\infty$ & 91.13 \\
            & Sliced (Finite) & Gaussian & Gaussian & 1024 & 91.13 \\
            & Sliced (Finite) & Gaussian & Laplace & 1024 & 90.25 \\
        \midrule
        \multirow{7}{*}{\textbf{KSD}} 
            & Unsliced & Gaussian & Gaussian & -- & 91.31 \\
            & Unsliced & Gaussian & Laplace & -- & 91.18  \\
            & Unsliced & IMQ & Gaussian & -- & \textbf{91.90}\\
            & Unsliced & IMQ & Laplace & -- & 91.12 \\
            & Sliced (Analytic) & Gaussian & Gaussian & $\infty$ & 91.11 \\
            & Sliced (Finite) & Gaussian & Gaussian & 1024 & 91.36  \\
            & Sliced (Finite) & Gaussian & Laplace & 1024 & 90.70\\
        \bottomrule
    \end{tabular}
    \label{tab:ablation-results}
\end{table}

\textbf{Strong visual representations.} We evaluate the benefit of the analytic discrepancy formulation in the self-distillation setting. KerJEPA achieves strong frozen-representation performance across varying model sizes, outperforming the corresponding DINO and iBOT baselines on both linear probing and $k$-NN classification (\cref{tab:imagenet-classification}) on ImageNet. Outside downstream classification, KerJEPA features also enable DINO-level semantic segmentation (see Appendix \ref{app:segm}); we believe this can further be improved with the inclusion of a masked image modeling loss, which we leave to future work.
\begin{table*}[h]
  \vspace{-\baselineskip}
  \centering
  \setlength{\tabcolsep}{10pt}
  \renewcommand{\arraystretch}{1.1}
    \centering
    \caption{\textbf{ImageNet-1k classification.} Top-1 accuracy (\%) on frozen $\clstoken$ features. KerJEPA out-competes baselines and shows favorable scaling across model sizes.}
    \label{tab:imagenet-classification}
    \small
    \begin{tabular}{cccc}
      \toprule
      Backbone & Method & k-NN & Linear \\
      \midrule
      \midrule
      \multirow[c]{3}{*}{ViT-S/16}
        & DINO    & 74.45 & 76.17 \\
        & iBOT    & 75.15 & 77.08 \\
        & KerJEPA\textsubscript{\tiny $\cwmmd$} & \textbf{76.09} & \textbf{77.85} \\
        & KerJEPA\textsubscript{\tiny $\cwksd$} & 75.40 & 77.65 \\
      \midrule
      \multirow[c]{3}{*}{ViT-B/16}
        & DINO    & 76.10 & 77.99 \\
        & iBOT    & 77.02 & 79.37 \\
        & KerJEPA\textsubscript{\tiny $\cwmmd$} & \textbf{78.95} & \textbf{80.80} \\
        & KerJEPA\textsubscript{\tiny $\cwksd$} & 78.58 & 80.65 \\
      \midrule
      \multirow[c]{2}{*}{ViT-L/16}
        & iBOT    & 77.94 & 80.87 \\
        & KerJEPA\textsubscript{\tiny $\cwmmd$} & \textbf{80.01} & 81.47 \\
        & KerJEPA\textsubscript{\tiny $\cwksd$} & 79.06 & \textbf{81.79} \\
      \bottomrule
    \end{tabular}
\end{table*}

\textbf{Stability and convergence.} Training in a self-distillation setting requires careful coordination of the optimization dynamics.
We observe that the learning rate and weight decay schedules are closely coupled.
Contrary to other methods, learning rate should not be decayed to zero, as doing so can substantially degrade representation quality, and in some cases lead to collapse. The terminal weight decay must likewise be chosen in conjunction with this minimum learning rate and that higher values of decay allowed for higher final learning rates. To maintain stability, layerwise learning-rate decay, patch scaling, projector batch normalization, and target standardization were beneficial. Across multiple runs with different random seeds, models trained without target standardization collapse, whereas applying standardization prevents these failures.

\section{Discussion}
\label{sec:discussion}
This paper unifies a design space for distributional regularizers for Euclidean SSL. Our analysis and results clarify the trade-offs between computation, stability, and representation geometry in practice.

\textbf{Computational tradeoffs.} While the analytically sliced variations of MMD and KSD are robust and exhibit accelerated convergence in certain scenarios, they formally incur quadratic computational complexity and rely on large-dimension asymptotes. In regimes where batch sizes do not exceed the total number of slices, the total benefit of slicing is not realized, barring any cross-device communication overhead in distributed settings.
Given that these algorithms require extended training schedules, the stability of exact (or analytically sliced) methods may outweigh the theoretical speedups of finite slicing.
Our theoretical results shed light on the nature of the kernels induced by slicing, particularly their heavy-tailed and dimension-dependent nature, which allows us to stabilize training and achieve faster convergence.
For larger-batch settings, computation can be reduced using Random Fourier Features \citep{rff, random-stein-feats, mmd-rff}, high-fidelity MMD and KSD coresets \citep{compress-test, debiased-dist-compress}, or B-statistic estimators \citep{bstat} that aggregate pairwise computations across distributed devices. 
Our experimental results show that slicing, while asymptotically more efficient in the limit of large batch size, is less stable and slower to converge on shorter training horizons.

\textbf{Representation geometry.} A critical consideration for these methods is their interaction with standard SSL architectures. Training employs a backbone-projector pair, where geometry is imposed on the projector's output, yet inference relies solely on the backbone \citep{guillotine}. We view the projector as a non-linear pushforward map; this decoupling obscures the geometry induced at the bottleneck.  Experimentally, we observe that the choice between Gaussian and Laplace priors on non-sliced experiments had negligible impact on downstream classification performance. Consequently, it remains unclear whether performance gains stem from the specific choice of isotropic prior or simply from the favorable learning dynamics of Euclidean gradients.

\bibliography{conference}
\bibliographystyle{conference}

\pagebreak
\appendix
\section{Shift-Invariant Kernels}

Kernel methods provide a powerful framework for comparing probability distributions by embedding them into a \emph{Reproducing Kernel Hilbert Space} (RKHS) \citep{rkhs}.
Let $k:\mathbb{R}^d \times \mathbb{R}^d \to \mathbb{R}$ be a positive definite kernel with associated RKHS $\mathcal{H}$.
The kernel defines a feature map $\phi: \mathbb{R}^d \to \mathcal{H}$ such that $k(x,x') = \langle \phi(x), \phi(x') \rangle_{\mathcal{H}}$. For any probability distribution $\prob{P}$ on $\mathbb{R}^d$, its \emph{kernel mean embedding} is defined as
\begin{equation*}
\mu_{\prob{P}} = \mathbb{E}_{\rv{x} \sim \prob{P}}[k(\rv{x}, \cdot)] = \mathbb{E}_{\rv{x} \sim \prob{P}}[\phi(\rv{x})] \in \mathcal{H}.
\label{eqn:mean-embedding}
\end{equation*}
This embedding maps a distribution into the RKHS, allowing us to measure distances between distributions via their embeddings. If the kernel $k$ is \emph{characteristic}, this mapping is injective: $\mu_{\prob{P}} = \mu_{\prob{Q}}$ if and only if $\prob{P} = \prob{Q}$ \citep{metrics-prob-measures}. This property ensures that kernel-based discrepancies capture full distributional differences, not merely differences in low-order moments.

Often, one employs a continuous, shift-invariant kernel $k(x,y) = k(x-y)$, which, by \emph{Bochner's Theorem} \citep{bochner},
admits a Fourier representation on non-negative Borel measures with a \emph{spectral density} $\rho_k$ defined as 
\begin{equation*}\label{eq:bochner}
k(x-y) = \int_{\mathbb{R}^d} \exp(\imag \omega^\top (x-y))  \rho_k(\omega)\dif\omega.
\end{equation*}
A common shift-invariant kernel of interest, categorized by its fast decay profile is the Gaussian Kernel
\begin{equation*}
\kgsn(x,y) = \exp(-\gamma \|x-y\|^2_2), 
\qquad 
\rho_\mathrm{gsn}(\omega) = \frac{1}{(4\pi\gamma)^{d/2}} \exp\left(-\frac{1}{4\gamma}\|\omega\|^2_2\right).
\label{eqn:gaussian-rbf}
\end{equation*}
Here, the bandwidth $\gamma > 0$ controls smoothness between global and local dependencies between samples.

We draw attention to a family of kernels of interest, namely, the \emph{Inverse Multiquadric Kernel} (IMQ). Unlike the Gaussian kernel, which has exponentially decaying tails, the IMQ decays at a polynomial rate (heavy-tailed) and is defined as
\begin{equation*}
\kimq(x,y) = \left( 1 + \alpha\|x-y\|^2_2 \right)^{-\beta}, 
\qquad
\rho_\mathrm{imq}(\omega) = \frac{2^{1-\beta}}{\Gamma(\beta)(2\pi\alpha)^{d/2}} \left( \frac{\|\omega\|_2}{2\sqrt{\alpha}} \right)^{\nu} K_{\nu}\left( \frac{\|\omega\|_2}{\sqrt{\alpha}} \right),
\label{eqn:imq-kernel}
\end{equation*}
where $\Gamma$ is the Gamma function and $K_{\nu}$ is the modified Bessel of second kind, order $\nu$. The parameter $\beta > 0$ acts as a shape parameter that determines the heaviness of the tails (rate of polynomial decay). $\alpha > 0$ serves as the inverse length scale. The IMQ kernel can be interpreted as an infinite scale mixture of Gaussian kernels. This heavy-tailed property allows it to capture long-range dependencies that are often missed by the fixed-scale Gaussian kernel. 
\section{Isotropic Prior Distributions}
\label{app:priors}

Representations are shaped using a selected prior distribution $p(x)$ as a learning target. Different isotropic priors induce different tail geometries. We incorporate these priors in different ways, either by taking expectations over $p(x)$ or by matching it's structure via the score $\nabla_x \log p(x)$ and characteristic function $\phi(\omega)$, all without having to drawing samples.

The characteristic function (CF) is defined the Fourier transform of the density with frequencies $\omega\in\mathbb{R}^d$ 
\begin{equation*}
\phi_{\prob{P}}(\omega)
=
\mathbb{E}_{\rv{x}\sim\prob{P}}
\left[
\exp\left(\imag \omega^\top \rv{x}\right)
\right].
\label{eq:cf}
\end{equation*}
Given samples \(\rv{x}_1,\ldots,\rv{x}_n\overset{\mathrm{iid}}{\sim}\prob{P}\), and the empirical distribution \(\empprob{P}_n=n^{-1}\sum_{i=1}^n\delta_{\rv{x}_i}\), we can write the empirical characteristic function (ECF) and its score-weighted counterpart as 
\begin{equation*}
\widehat{\phi}_n(\omega)
:=
\frac{1}{n}\sum_{i=1}^{n}
\exp(\iu\omega^\top\rv{x}_i),
\qquad
\widehat{\phi}_{s_{\prob{Q}},n}(\omega)
:=
\frac{1}{n}\sum_{i=1}^{n}
s_{\prob{Q}}(\rv{x}_i)\exp(\iu\omega^\top\rv{x}_i).
\label{eqn:empirical-spectral-features}
\end{equation*}
which enables direct matching between the ECF to a known CF. We provide examples for three isotropic priors that can be used a as regularizer below.

\begin{table}[h!]
    \caption{Isotropic prior distributions: probability density $p(x)$, scores $\nabla_x \log p(x)$, and their characteristic function $\phi(\omega)$\label{table:priors} used to construct different discrepancy objectives.}
    \centering
    \resizebox{\textwidth}{!}{%
    \footnotesize
    \begin{tabular}{l c c c c}
        \toprule
        \textbf{Distribution} & \textbf{Notation} & $p(x)$ & $\nabla_x \log p(x)$ & $\phi(\omega)$ \\
        \midrule
        \midrule

        \textbf{Gaussian} & 
        $\mathcal{N}(0, \sigma^2 I_d)$ & 
        $\displaystyle \frac{1}{(2\pi\sigma^2)^{d/2}} \exp\left(-\frac{1}{2\sigma^2}\|x\|^2_2\right)$ & 
        $\displaystyle -\frac{1}{\sigma^2}x$ & 
        $\displaystyle \exp\left(-\frac{1}{2}\sigma^2 \|\omega\|^2_2\right)$ \\
        \midrule

        \textbf{Laplace} & 
        $\textrm{Lap}(0, \sigma I_d)$ & 
        $\displaystyle \frac{\Gamma(d/2)}{2\pi^{d/2}\sigma^d \Gamma(d)} \exp\left(-\frac{1}{\sigma}\|x\|_2\right)$ & 
        $\displaystyle -\frac{1}{\sigma}\frac{x}{\|x\|_2}$ & 
        $\displaystyle \left(1 + \sigma^2 \|\omega\|^2_2\right)^{-\frac{d+1}{2}}$ \\
        \midrule

        \textbf{Student-t} & 
        $\mathcal{T}_\nu(0, \sigma^2 I_d)$ & 
        $\displaystyle \frac{\Gamma(\frac{\nu+d}{2})}{\Gamma(\frac{\nu}{2})(\nu\pi\sigma^2)^{d/2}} \left(1+\frac{1}{\nu\sigma^2}\|x\|^2_2\right)^{-\frac{\nu+d}{2}}$ & 
        $\displaystyle -\frac{\nu+d}{\nu\sigma^2 + \|x\|^2_2} x$ & 
        $\displaystyle \frac{K_{\nu/2}(\sqrt{\nu}\sigma\|\omega\|_2) (\sqrt{\nu}\sigma\|\omega\|_2)^{\nu/2}}{\Gamma(\nu/2) 2^{\nu/2-1}}$ \\
        
        \bottomrule
    \end{tabular}%
    }
\end{table}
\section{Kernel Discrepancies}
\label{app:discrepancies}
In this section, we derive variations of the MMD and KSD with corresponding sample-based estimators, discussing the features and difficulties associated with different choices of kernels and priors. We note that all discrepancies that can be written under Bochner's theorem can be approximated via slicing paired with 1D quadrature or via RFF.

\subsection{Maximum Mean Discrepancy}
\label{app:mmd}

Restating the definition, the \emph{Maximum Mean Discrepancy} (MMD) is an RKHS distance measure between probability distributions $\prob{P}$ and $\prob{Q}$ defined via kernel mean embeddings from each distribution. For a kernel $k$ associated with RKHS $\set{H}$, it is defined as
\begin{equation*}
\mmd[k](\prob{P},\prob{Q}) = \|\mu_{\learnprob} - \mu_{\targetprob}\|_{\set{H}}.
\end{equation*}
Equivalently, the square of this metric has the following useful form:
\begin{equation*}
\mmd[k]^2(\prob{P}, \prob{Q})
    =
\mathbb{E}_{\rv{x},\rv{x}' \sim \prob{P}}[k(\rv{x},\rv{x}')] 
+ \mathbb{E}_{\rv{y},\rv{y}' \sim \prob{Q}}[k(\rv{y},\rv{y}')] 
- 2\mathbb{E}_{\rv{x} \sim \prob{P}, \rv{y} \sim \prob{Q}}[k(\rv{x},\rv{y})].
\label{eqn:mmd-expanded-appendix}
\end{equation*}
Altogether, we provide the following MMD estimators, with $n,m$ denoting the number of samples drawn from the distribution under test and the prior distribution respectively:
\begin{enumerate}
    \item $\approximate{\fnfont{UMMD}}_{k, n, m}^2$ \cref{eq:mmd:u-statistic}: a general unbiased estimator for MMDs with arbitrary kernels and priors;
    \item $\approximate{\fnfont{CMMD}}_{k, n}^2$ \cref{eq:mmd:bochner-statistic}: a consistent estimator for shift-invariant kernels $k$ and priors with known characteristic functions, which requires only $O(n)$ computational complexity; and
    \item $\approximate{\fnfont{UGMMD}}_{\sigma^2, n}^2$ \cref{eq:mmd:gaussian:u-statistic}: an unbiased estimator for Gaussian kernels with prior $\mathcal{N}(0,\sigma^2I_d)$, with reduced variance.
\end{enumerate}
The following U-statistic is an unbiased estimator \cite{kernel-two-sample} of the MMD, given $\rv{X}_i\overset{\mathrm{iid}}{\sim} \learnprob\quad\text{and}\quad \rv{Y}_i\overset{\mathrm{iid}}{\sim}\targetprob$ where
\begin{equation}
    \label{eq:mmd:u-statistic}
    \approximate{\fnfont{UMMD}}_{k, n, m}^2(\prob{P}, \prob{Q}) 
    = \frac{1}{n(n-1)}\sum_{i=1}^n\sum_{j\neq i}k(\rv{X}_i, \rv{X}_j) 
    + \frac{1}{m(m-1)}\sum_{i=1}^m\sum_{j\neq i}k(\rv{Y}_i, \rv{Y}_j)
    - 2\sum_{i=1}^n\sum_{j=1}^mk(\rv{X}_i, \rv{Y}_j).
\end{equation}
It is well-known that estimation of the MMD via this U-statistic achieves dimension-free sample complexity \cite{kernel-two-sample}. However, there are at least two qualities of this estimator one might hope to improve:
\begin{enumerate}
    \item It involves pairwise kernel evaluations, which scales quadratically with the number of samples drawn from the distributions under comparison; and
    \item If one could compute the expectations over the prior ($\targetprob$) in closed form, variance can be reduced.
\end{enumerate}

With regard to the first point, we can turn to the Bochner integral formulation of the MMD, which would scale linearly with the number of samples. Particularly, we derive the CF MMD estimator (a V-estimator) below:
\begin{talign}
    \label{eq:mmd:bochner-statistic}
    \approximate{\fnfont{CMMD}}_{k, n}^2(\prob{P}, \prob{Q})
    = \int_{\mathbb{R}^d}\left\lvert\phi_{\empprob{P}^n}(\omega) - \phi_{\prob{Q}}(\omega)\right\rvert^2\rho_k(\omega)\dif\omega,\quad \empprob{P}_n = \frac{1}{n}\sum_{i=1}^n\delta_{\rv{x}_i},\quad \rv{x}_i\overset{\mathrm{iid}}{\sim}\prob{P},
\end{talign}
where $\phi_{\prob{P}}$ denotes the CF of the probability measure $\prob{P}$. 
Notably, under this estimator, there are no pairwise comparisons being made, and the complexity scales only as $O(n)$.
Here, we assume that $\phi_{\targetprob}$ is known analytically, which is often the case for priors used in practice (e.g. Gaussian, Laplace, or Student-t priors).
Note, however, that certain kernels (e.g. Student-t, IMQ) require their own quadrature for estimating their spectral density, adding additional overhead, which also scales with $d$.

With regard to the second criterion, computing the expectation over $\targetprob$ in closed form is generally intractable.
Exceptionally, this can be done in the scenario of Gaussian priors under Gaussian kernels.

\begin{proposition}\label{claim:mmd-gaussian-closed}
For any distribution $\prob{P}\in\probset{\mathbb{R}^d}$, we have
\begin{equation*}
    \label{eq:mmd-gaussian-closed}
    \begin{split}
        \mmd[\kgsn]^2(\prob{P}, \mathcal{N}(0,\sigma^2I_d))
        &= \mathbb{E}_{\rv{x}, \rv{x}'\sim \prob{P}}\exp(-\gamma\|\rv{x} - \rv{x}'\|_2^2) \\
        &\quad- \frac{2}{(1 + 2\gamma\sigma^2)^{d/2}}\mathbb{E}_{\rv{x}\sim \prob{P}}\exp\left(-\frac{\gamma}{1 + 2\gamma\sigma^2}\|\rv{x}\|^2_2\right) \\
        &\quad+ C^{\rm gauss}_{\gamma, \sigma, d},
    \end{split}
\end{equation*}
for a constant $C^{\rm gauss}_{\gamma, \sigma, d}$ independent of $\prob{P}$ given by
\begin{align*}
    C^{\rm gauss}_{\gamma, \sigma, d} := \frac{1}{(1 + 4\gamma\sigma^2)^{d/2}}.
\end{align*}
\end{proposition}

We prove this proposition \hyperref[proof:mmd-gaussian-closed]{below}.
First, we highlight the resulting unbiased U-statistic $\approximate{\fnfont{UGMMD}}_{\sigma^2, n}^2(\prob{P})$ in the case of a Gaussian prior with a Gaussian kernel---this is also known as the \emph{BHEP} estimator \cite{Baringhaus1988}:
\begin{equation}
    \label{eq:mmd:gaussian:u-statistic}
    \begin{split}
    \approximate{\fnfont{UGMMD}}_{\sigma^2, n}^2(\prob{P})
    &= \frac{1}{n(n-1)}\sum_{i=1}^n\sum_{j\neq i}\exp(-\gamma\|\rv{x}_i - \rv{x}_j\|_2^2)\\
    &\quad- \frac{2}{(1 + 2\gamma\sigma^2)^{d/2}}\sum_{i=1}^n\exp\left(-\frac{\gamma}{1 + 2\gamma\sigma^2}\|\rv{x}_i\|^2_2\right) \\
    &\quad+ C^{\rm gauss}_{\gamma, \sigma, d},
    \end{split}
\end{equation}
where $\{\rv{x}_i\}_{i=1}^n\overset{\mathrm{iid}}{\sim}\prob{P}$, and where $C^{\rm gauss}_{\gamma,\sigma,d}$ is independent of $\prob{P}$ and given in Proposition \cref{claim:mmd-gaussian-closed}.

\begin{proof}
    \label{proof:mmd-gaussian-closed}
    \emph{Proof of Proposition \cref{claim:mmd-gaussian-closed}.}
    Let $\gaussshorthand = \mathcal{N}(0,\sigma^2 I_d)$.
    Starting from \eqref{eqn:mmd-expanded}, we have
    \begin{align*}
        \mmd[\kgsn]^2(\prob{P}, \gaussshorthand) = \mathbb{E}_{\rv{x}, \rv{x}'\sim \prob{P}}[\kgsn(\rv{x}, \rv{x}')] + \underbrace{\mathbb{E}_{\rv{y}, \rv{y}'\sim \gaussshorthand}[\kgsn(\rv{y}, \rv{y}')]}_{\rm A} - 2\underbrace{\mathbb{E}_{(\rv{x}, \rv{y})\sim \prob{P}\otimes \gaussshorthand}[k(\rv{x}, \rv{y})]}_{\rm B}.
    \end{align*}
    We simplify terms $\rm A$ and $\rm B$ individually. Starting with $\rm A$,
    \begin{align*}
    \mathrm{A} &\defeq \mathbb{E}_{\rv{y},\rv{y}' \sim \targetprob}[\kgsn(\rv{y},\rv{y}')]\\
    &= \int_{\mathbb{R}^d}\int_{\mathbb{R}^d} \exp\left(-\gamma \|y-y'\|^2_2\right) \frac{1}{(2\pi \sigma^2)^d} \exp\left(-\frac{1}{2\sigma^2}(\|y\|^2_2 + \|y'\|^2_2)\right)  \dif y  \dif y'\\
    &= \frac{1}{(1 + 4 \gamma \sigma^2)^{d/2}} \defeq  C^{\rm gauss}_{\gamma,\sigma,d}.
    \end{align*}
    Now, for $\mathrm{B}$, we have
    \begin{talign*}
        \mathbb{E}_{\rv{y} \sim \gaussshorthand}[\kgsn(x,\rv{Y})]
        &= \int_{\mathbb{R}^d} \exp(-\gamma \|x-y\|^2_2) \frac{1}{(2\pi\sigma^2)^{d/2}} \exp\left(-\frac{1}{2\sigma^2} \|y\|^2_2\right) \dif y.
    \end{talign*}
    This is a convolution of Gaussians, which evaluates to
    \begin{talign*}
        \mathbb{E}_{\rv{y} \sim \gaussshorthand}[\kgsn(x,\rv{y})]
        &= \frac{1}{(2\gamma\sigma^2 + 1)^{d/2}}\exp\left(-\frac{\gamma}{2\gamma\sigma^2 + 1}\|x\|^2_2\right).
    \end{talign*}
    Substituting into $\rm B$, we have
    \begin{talign*}
    \mathrm{B} &\defeq  \mathbb{E}_{\rv{x}\sim \prob{P}}\mathbb{E}_{\rv{y}\sim \gaussshorthand}[\kgsn(\rv{x}, \rv{y})]
    =\frac{1}{(2\gamma\sigma^2 + 1)^{d/2}}\mathbb{E}_{\rv{x}\sim \prob{P}}\left[\exp\left(-\frac{\gamma}{2\gamma\sigma^2 + 1}\|\rv{x}\|^2_2\right)\right].
    \end{talign*}
    Substituting $\rm A$ and $\rm B$ above completes the proof.
\end{proof}

\subsection{Kernel Stein Discrepancy}
\label{app:ksd}

Restating the definition, the \emph{Kernel Stein Discrepancy} (KSD) \cite{kernel-gof, ksd, sample-quality-kernels, ksdd} is an RKHS-based measure of discrepancy between a probability distribution $\prob{P}$ and a target distribution $\targetprob$. For an absolutely continuous target distribution $\targetprob$, let $s_{\targetprob}(x) = \nabla_x \log \targetprob(x)$ denote its score function. The Langevin Stein operator $\opfont{A}_{\targetprob}$ acts on a vector-valued test function $f:\mathbb{R}^d \to \mathbb{R}^d$ according to
\begin{equation*}
    (\opfont{A}_{\targetprob}f)(x) = s_{\targetprob}(x)^\top f(x) + \nabla_x \cdot f(x).
\end{equation*}
Under suitable regularity conditions, Stein's identity states that $\mathbb{E}_{x\sim\targetprob}[(\opfont{A}_{\targetprob}f)(x)] = 0$. Consequently, deviations from this identity under $\prob{P}$ provide a measure of discrepancy from $\targetprob$. Taking the supremum over the unit ball of a vector-valued RKHS and applying the reproducing property yields the squared KSD:
\begin{equation*}
    \ksd^2(\prob{P}, \targetprob) = \mathbb{E}_{x,x'\sim\prob{P}}[k_\mathrm{stein}(x,x')],
\end{equation*}
where the Stein kernel is defined by applying the Stein operator to both arguments of a positive-definite base kernel $k:\mathbb{R}^d \times \mathbb{R}^d \to \mathbb{R}$:
\begin{talign*}
    k_\mathrm{stein}(x,y) =
    s_{\targetprob}(x)^\top k(x,y)s_{\targetprob}(y)
    + s_{\targetprob}(x)^\top\nabla_y k(x,y)
    + \nabla_x k(x,y)^\top s_{\targetprob}(y)
    + \operatorname{tr}\big(\nabla_x\nabla_y^\top k(x,y)\big).
\end{talign*}
Importantly, the target distribution enters only through its score function, eliminating the need to sample from $\targetprob$ or evaluate its normalization constant. This distinguishes KSD from the general MMD formulation, which typically requires samples from both distributions or closed-form evaluations of kernel expectations under the target. The resulting discrepancy can therefore be interpreted as a target-dependent kernel discrepancy in which the base kernel is transformed according to the geometry induced by $\targetprob$.

We consider variations of the KSD induced by different base kernels and target isotropic priors. Let $\{\rv{x}_1,\ldots,\rv{x}_n\}\subset\mathbb{R}^d$ be independent samples from an arbitrary distribution $\prob{P}$, with empirical distribution $\empprob{P} = n^{-1}\sum_{i=1}^n\delta_{\rv{x}_i}$. An unbiased U-statistic estimator of the squared KSD is given by
\begin{equation*}
    \approximate{\ksd}^2(\prob{P},\targetprob)
    =
    \frac{1}{n(n-1)}
    \sum_{i=1}^n\sum_{j\neq i}
    k_\mathrm{stein}(\rv{x}_i,\rv{x}_j).
\end{equation*}
We examine Gaussian, IMQ, and general shift-invariant kernels, together with the Gaussian, Laplace, and Student-$t$ isotropic priors in \cref{app:priors}. The required kernel gradients and mixed Hessian traces are provided in \cref{app:stein-grads}. Combining these derivatives with the corresponding prior score functions yields closed-form Stein kernels for each kernel--prior combination. For radial kernels, we use the identity $\nabla_x k(x,y)=-\nabla_y k(x,y)$ to simplify the resulting expressions.

The U-statistic above requires pairwise kernel evaluations and therefore scales quadratically with the number of samples. For shift-invariant kernels, however, Bochner's theorem provides an alternative spectral representation that avoids explicit pairwise comparisons. We write the corresponding spectral KSD estimator as a function of the ECF and score-weighted ECF \cref{eqn:empirical-spectral-features} as
\begin{equation*}
    \label{eq:ksd:bochner-statistic}
    \approximate{\fnfont{CKSD}}_{k,n}^2(\prob{P},\targetprob)
    =
    \int_{\mathbb{R}^d}
        \left\|
        \widehat{\phi}_{s_{\prob{Q}},n}(\omega)
        +\iu\omega\widehat{\phi}_n(\omega)
        \right\|^2
        \rho_k(\omega)\,\dd\omega.
\end{equation*}
Unlike the U-statistic, this estimator does not require pairwise kernel evaluations and can therefore be evaluated in $O(n)$ time for a fixed number of frequency evaluations. As with the spectral MMD estimator, the integral can be approximated using samples from the spectral density, providing a computationally efficient alternative for large batches. Because the score function is available in closed form for the isotropic priors considered in \cref{app:priors}, these kernel--prior combinations can be evaluated directly, yielding a flexible family of distributional regularizers without explicit representation normalization. Additionally, we may estimate the marginals along slices, allowing for the same quadrature estimation as done in SIGReg.

We provide a table with gradients and Hessians for the Gaussian and IMQ kernels needed to assemble the discrepancy, noting that the gradients for a radial kernel are odd ($\nabla_x k(x,y) = - \nabla_yk(x,y)$).
\begin{table}[h]
    \centering
    \small
    \setlength{\tabcolsep}{3pt}
    \caption{Gradients and Hessians for the Gaussian, IMQ, and general
    shift-invariant kernels..}
    \begin{tabularx}{\textwidth}{l
        >{\centering\arraybackslash}X
        >{\centering\arraybackslash}X}
        \toprule
        \textbf{$k(x,y)$} &
        \textbf{$\nabla_x k(x,y)$} &
        \textbf{$\operatorname{tr}(\nabla_x\nabla_y k(x,y))$} \\
        \midrule

        \textbf{Gaussian} &
        $\displaystyle
        -2\gamma(x-y)e^{-\gamma\|x-y\|_2^2}$ &
        $\displaystyle
        (2\gamma d-4\gamma^2\|x-y\|_2^2)
        e^{-\gamma\|x-y\|_2^2}$ \\

        \midrule

        \textbf{IMQ} &
        $\displaystyle
        -2\alpha\beta(x-y)
        (1+\alpha\|x-y\|_2^2)^{-(\beta+1)}$ &
        $\displaystyle
        4\alpha^2\beta(\beta+1)\|x-y\|_2^2
        (1+\alpha\|x-y\|_2^2)^{-(\beta+2)}$ \\

        \midrule

        \textbf{Bochner} &
        $\displaystyle
        \int_{\mathbb{R}^d}
        \imag\omega\,
        e^{\imag\omega^\top(x-y)}
        \rho(\omega)\,d\omega$ &
        $\displaystyle
        \int_{\mathbb{R}^d}
        \|\omega\|_2^2
        e^{\imag\omega^\top(x-y)}
        \rho(\omega)\,d\omega$ \\

        \bottomrule
    \end{tabularx}
    \label{app:stein-grads}
\end{table}

\subsection{Sliced Regularization Algorithms}
We provide two examples of sliced discrepancy regularization using both MMD and KSD. The MMD setting generalizes SIGReg and supports different kernels and priors

This process decomposes the computational cost into two efficient steps: projections of $n$ embeddings of dimension $d$ onto a set of $r$ random directions, and quadratures to estimate the spectral integral using a fixed set of $u$ integration knots. 
This reduces the complexity from $\Theta(n^{2}d)$ pairwise kernel evaluations to $\Theta(nr(d+u))$, making it linear in the batch size. 
These fast quadrature-based methods are primarily applicable when using the Gaussian or IMQ kernels combined with Gaussian, Laplace, and Student-t priors, as their spectral density, projected marginals, and CFs are analytic and well-behaved. In the case of the IMQ kernel and Student-t CF, Bessel functions only need to be computed once on their one-dimensional marginals at the known quadrature knots on initialization. 
We implement the two sliced methods, termed the sliced MMDReg and sliced KSDReg, as shown in \cref{app:kerjepa-pseudo}. and we restrict our implementation to Gaussian kernel with Gaussian and Laplace priors for simplicity, as the Gaussian kernel as weighting function allows us to approximate the integral with (symmetric) Gauss-Hermite quadratures. 

\begin{figure}[h]
    \centering
    \begin{minipage}{0.48\textwidth}
\begin{lstlisting}[language=Python, basicstyle=\ttfamily\tiny, frame=single, title={Sliced MMD as generalized SIGReg}]
class SlicedMMReg(Module):
    def __init__(self, n_knots):
        super().__init__()
        # Full quadrature (no symmetry assumption)
        k, w = polynomial.hermite.hermgauss(n_knots)
        self.knots = tensor(k * sqrt(2)).view(1,1,-1)
        self.weights = tensor(w / sqrt(pi)).view(1,1,-1)

    def target_cf(self, w, sigma, prior):
        if prior == 'gaussian':
            return exp(-0.5 * (sigma**2) * w**2)
        elif prior == 'laplace':
            return 1.0 / (1.0 + (sigma**2) * w**2)

    def forward(self, z, slices, sigma, prior):
        # Sample slices and project
        noise = randn(slices, z.size(1))
        theta = normalize(noise, dim=1)
        proj = (z @ theta.T).unsqueeze(-1)


        # MSE ECF and CF
        args = proj * self.knots
        emp_cf = mean(torch.cos(args), dim=0)
        tgt_cf = self.target_cf(self.knots, sigma, prior)
        err = (emp_cf - tgt_cf.squeeze()).pow(2)


        # Standard integration
        weights = self.weights
        smmd = sum(err * weights, dim=-1).mean()
        
        return smmd
\end{lstlisting}
    \end{minipage}
    \hfill
    \begin{minipage}{0.48\textwidth}
\begin{lstlisting}[language=Python, basicstyle=\ttfamily\tiny, frame=single, title={Sliced KSD as a Stein analogue to LeJEPA}]
class SlicedKSDReg(Module):
    def __init__(self, n_knots):
        super().__init__()
        # Full quadrature (no symmetry assumption)
        k, w = polynomial.hermite.hermgauss(n_knots)
        self.knots = tensor(k * sqrt(2)).view(1,1,-1)
        self.weights = tensor(w / sqrt(pi)).view(1,1,-1)

    def score(self, z, sigma, prior):
        if prior == 'gaussian':
            return -z / sigma**2
        elif prior == 'laplace':
            return -normalize(z) / sigma

    def forward(self, z, slices, sigma, prior):
        # Sample slices and project
        noise = randn(slices, z.size(1))
        theta = normalize(noise, dim=1)
        proj = z @ theta.T
        s = self.score(proj, sigma, prior) @ theta.T
        
        # MSE Stein ECF
        args = proj * self.knots
        w = self.knots.expand_as(args)
        real = s * cos(args) - w * sin(args)
        imag = s * sin(args) + w * cos(args)
        disc = mean(real,0).pow(2) + mean(imag,0).pow(2)

        # Standard integration
        weights = self.weights
        sksd = sum(disc * weights, dim=-1).mean()
        
        return sksd
\end{lstlisting}
    \end{minipage}
    \caption{Pseudo-code for the finite-sliced discrepancy regularization losses for both Gaussian and Laplace priors using the Gaussian kernel. Integration is done as Gauss-Hermite quadrature approximation. Other choice of kernel may require additional quadrature estimation}
    \label{app:kerjepa-pseudo}
\end{figure}

\section{Semantic Segmentation with KerJEPA}\label{app:segm}
In addition to providing exceptionally strong features for downstream classification tasks, we sought out to test whether KerJEPA features also enable accurate semantic segmentation.
Towards this end, we evaluated semantic segmentation on the ADE20K dataset \citep{Zhou_2017_CVPR} via linear probing, and compare our results to those derived from iBOT and DINO features.
The results are given in \cref{tab:ade20k-segmentation}.

\begin{table*}[h]
    \centering
    \caption{\textbf{Semantic segmentation on ADE20K.} mIoU/mAcc on frozen patch features. iBOT shows strongest performance but is trained with masked objectives.}
    \label{tab:ade20k-segmentation}
    \begin{tabular}{cccc}
      \toprule
      Backbone & Method & mIoU & mAcc \\
      \midrule
      \midrule
      \multirow[c]{4}{*}{ViT-S/16}
        & \textcolor{gray}{iBOT}    & \textcolor{gray}{\textbf{33.01}} & \textcolor{gray}{\textbf{42.34}} \\\arrayrulecolor{gray}\cline{2-4}\arrayrulecolor{black}
        & DINO    & 30.29 & 39.05 \\
        & KerJEPA\textsubscript{\tiny $\cwmmd$}& 30.67 & 39.05 \\
        & KerJEPA\textsubscript{\tiny $\cwksd$}     & 30.54 & 38.76 \\
      \midrule
      \multirow[c]{4}{*}{ViT-B/16}
        & \textcolor{gray}{iBOT}     & \textcolor{gray}{\textbf{38.08}} & \textcolor{gray}{\textbf{47.41}} \\\arrayrulecolor{gray}\cline{2-4}\arrayrulecolor{black}
        & DINO     & 34.77 & 43.93 \\
        & KerJEPA\textsubscript{\tiny $\cwmmd$}  & 34.52 & 43.42 \\
        & KerJEPA\textsubscript{\tiny $\cwksd$}  & 35.52 & 44.58 \\
      \midrule
      \multirow[c]{3}{*}{ViT-L/16}
        & \textcolor{gray}{iBOT}     & \textcolor{gray}{\textbf{41.02}} & \textcolor{gray}{\textbf{50.86}} \\\arrayrulecolor{gray}\cline{2-4}\arrayrulecolor{black}
        & KerJEPA\textsubscript{\tiny $\cwmmd$} & 37.10 & 46.22 \\
        & KerJEPA\textsubscript{\tiny $\cwksd$} & 37.27 & 46.29 \\
      \bottomrule
    \end{tabular}
\end{table*}

We observe that iBOT remains ahead in the context of semantic segmentation, but note that it leverages pretraining techniques, particularly a masked image modeling objective, that have been developed particularly for such a task.
That said, while KerJEPA features do not match iBOT's for the purpose of semantic segmentation, they are very competitive with DINO features towards this end, which itself is a strong baseline.

Altogether, the addition of a masked image modeling objective is largely orthogonal to the developments made in this paper, which mainly deal with the regularizer.
The results in \cref{tab:ade20k-segmentation} suggest that, when augmented with a masked image modeling objective during pretraining, KerJEPA can be very well suited to semantic segmentation as well.
Since our main focus in this work was to generalize the framework introduced by LeJEPA, our study focused largely on the simplest extension of LeJEPA, but such a masked image modeling extension presents an exciting avenue for future work.

\section{DAVIS Video Instance Segmentation}\label{app:davis}
\begin{figure}[h]
    \centering
    \begin{subfigure}[b]{0.32\linewidth}
        \centering
        \includegraphics[trim=0 .1cm 0 0,clip,width=\linewidth]{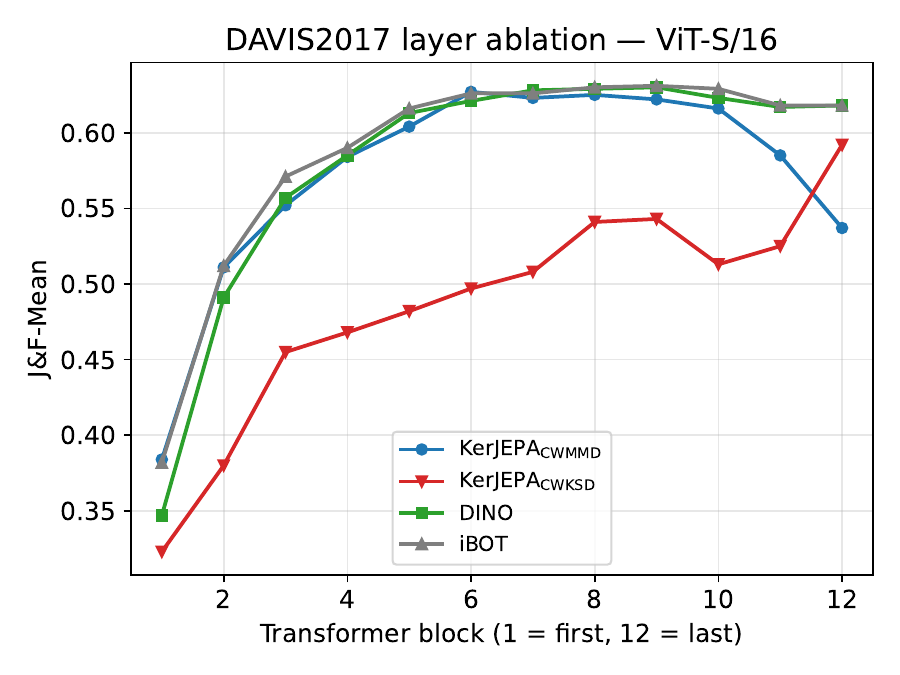}
        \caption{ViT-S}
        \label{fig:davis-vits}
    \end{subfigure}
    \hfill
    \begin{subfigure}[b]{0.32\linewidth}
        \centering
        \includegraphics[trim=0 .1cm 0 0,clip,width=\linewidth]{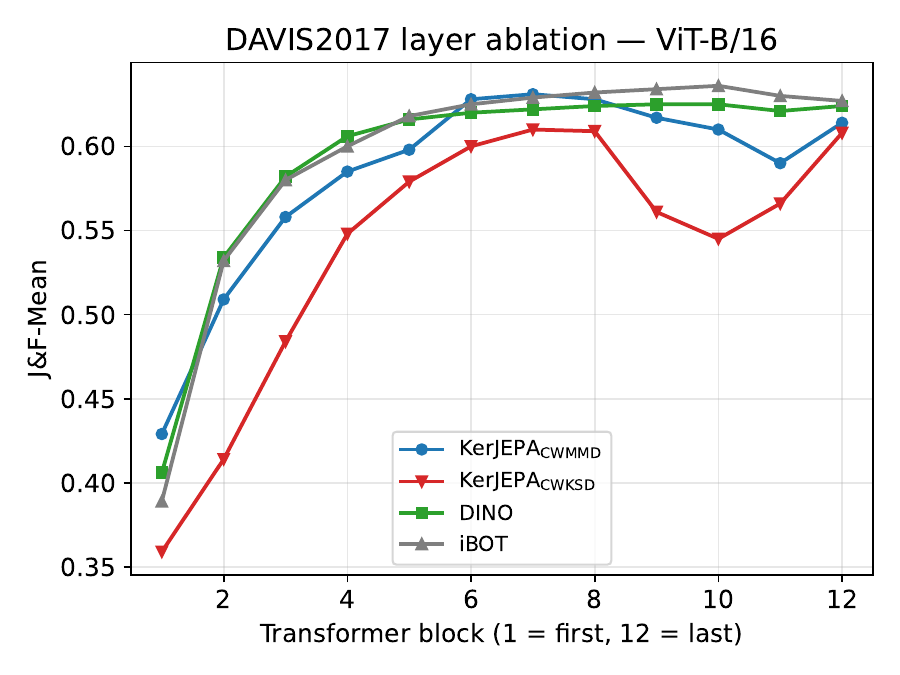}
        \caption{ViT-B}
        \label{fig:davis-vitb}
    \end{subfigure}
    \hfill
    \begin{subfigure}[b]{0.32\linewidth}
        \centering
        \includegraphics[trim=0 .1cm 0 0,clip,width=\linewidth]{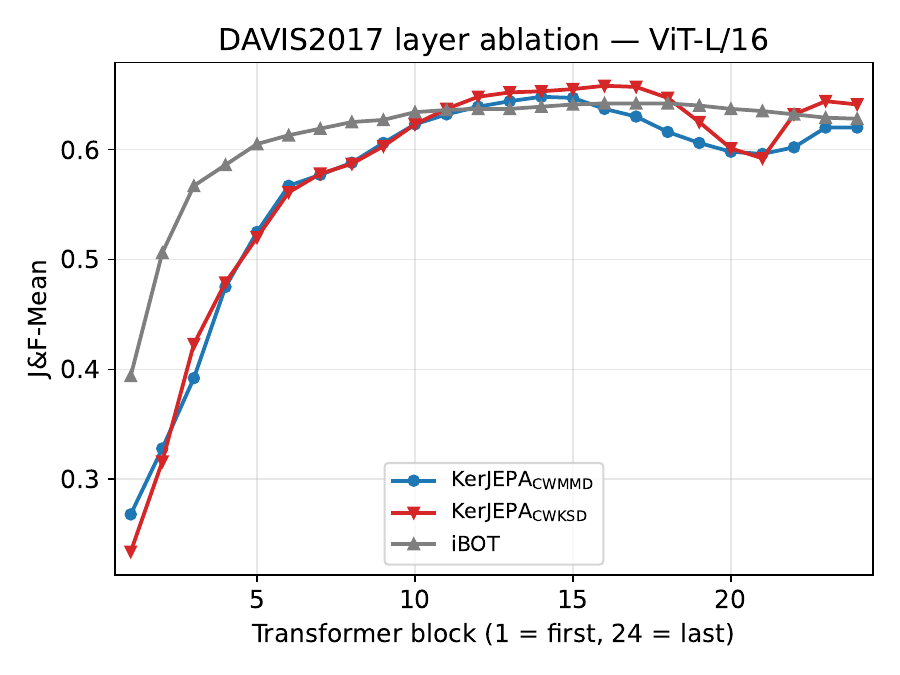}
        \caption{ViT-L}
        \label{fig:davis-vitl}
    \end{subfigure}
    \caption{DAVIS layer ablation across model sizes. Despite KerJEPA having weaker spatial features, they show favorable scale, with top performance even surpassing iBOT at intermediate layers without direct patch supervision.}
    \label{fig:davis-layer-ablation}
\end{figure}

We also examine video instance segmentation on the DAVIS-2017 benchmark \cite{pont20172017}, using the same evaluation protocol as DINO \cite{caron2021emerging}. We evaluate segmentation performance at each layer across each model size and compare our results to those from iBOT and DINO models. Results are shown in Figure \ref{fig:davis-layer-ablation}.

\section{Proofs of Theoretical Claims}\label{app:proofs}
\kummermmd*
\begin{proof}
    \label{proof:kummermmd}
    As we saw in \S\ref{sec:lejepa}, it holds that
    \begin{align*}
        \ep(\learnprob) = n\mmd[\kgsn]^2(\learnprob, \mathcal{N}(0, \sigma^2)).
    \end{align*}
    Let $\rv{x}\sim \learnprob$ and $\rv{N}\sim\mathcal{N}(0, \sigma^2I_d)$. Then, by the linearity of the expectation, we have
    \begin{talign*}
        \sigreg(\learnprob)
        &= \mathbb{E}_{\rv{\direction}\sim\mathcal{U}(\mathbb{S}^{d-1})}\left[\mmd[\kgsn]^2(\rv{\direction}^\top_\#\learnprob, \rv{\direction}^\top_\#\mathcal{N}(0, \sigma^2I_d))\right]\\
        &= \mathbb{E}_{\rv{x}, \rv{x}'\sim \learnprob}\mathbb{E}_{\rv{\direction}\sim\mathcal{U}(\mathbb{S}^{d-1})}\kgsn(\rv{\direction}^\top \rv{x}, \rv{\direction}^\top \rv{x}') \\
        &\quad+\mathbb{E}_{\rv{y}, \rv{y}'\sim \mathcal{N}(0, \sigma^2I_d)}\mathbb{E}_{\rv{\direction}\sim\mathcal{U}(\mathbb{S}^{d-1})}\kgsn(\rv{\direction}^\top \rv{y}, \rv{\direction}^\top \rv{y}')\\
        &\quad-
        2\mathbb{E}_{\rv{x}\sim \learnprob, \rv{y}\sim \mathcal{N}(0, \sigma^2I_d)}\mathbb{E}_{\rv{\direction}\sim\mathcal{U}(\mathbb{S}^{d-1})}\kgsn(\rv{\direction}^\top \rv{x}, \rv{\direction}^\top \rv{y}).
    \end{talign*}

    Moreover, by \citet[Corollary S3]{nadjahi2020statistical}, it holds that $(x, y)\mapsto\mathbb{E}_{\rv{\direction}\sim\mathcal{U}(\mathbb{S}^{d-1})}\kgsn(\rv{\direction}^\top x, \rv{\direction}^\top y) = \kummer(1/2; d/2; \gamma\|x - y\|_2^2)$ is a particular solution to \emph{Kummer's equation}; that is, $\kummer(a, b, z)$ is a solution to
    \begin{align*}
        z\frac{\dd^2 w}{\dd z^2} + (b - z)\frac{\dd w}{\dd z} - \alpha w = 0.
    \end{align*}

    Thus, substituting, we have
    \begin{align*}
        \sigreg(\learnprob)
        &= \mathbb{E}_{\rv{x}, \rv{x}'\sim \learnprob}\left[\kummer\left(\frac{1}{2}; \frac{d}{2}; \gamma\|\rv{x} - \rv{x}'\|_2^2\right)\right] +
        \mathbb{E}_{\rv{Y}, \rv{Y}'\sim \mathcal{N}(0, \sigma^2I_d)}\left[\kummer\left(\frac{1}{2};\frac{d}{2}; \gamma\|\rv{Y} - \rv{Y}'\|_2^2\right)\right]\\&\quad-
        2\mathbb{E}_{\rv{x}\sim P, \rv{y}\sim \mathcal{N}(0, \sigma^2 I_d)}\left[\kummer\left(\frac{1}{2}; \frac{d}{2}; \gamma\|\rv{x} - \rv{y}\|_2^2\right)\right]\\
        &\equiv \mmd[\kummerkernel]^2(\learnprob, \mathcal{N}(0,\sigma^2I_d)).
    \end{align*}

    This proves the first claim.
    Next, it is known \cite[Chapter 13]{abramowitz1948handbook} that for fixed $d$,
    \begin{align*}
        &\kummer\left(\frac{1}{2}; \frac{d}{2}; -\gamma\|x - y\|_2^2\right)\\
        &\quad\sim
        \Gamma\left(\frac{d}{2}\right)\left[\frac{1}{\Gamma(1/2)}\exp(-\gamma\|x-y\|_2^2)\left(-\gamma\|x-y\|_2^2\right)^{(1-d)/2} + \frac{1}{\Gamma((d-1)/2)}\gamma^{-1/2}\|x-y\|_2^{-1}\right],
    \end{align*}

    where $f(\xi)\sim g(\xi)$ denotes that $f$ is asymptotically equivalent to $g$ in $\xi$, in the sense that $\lim_{\xi\to\infty}f(\xi)/g(\xi) = 1$; the expression above is stated asymptotically for $\|x-y\|_2$. Thus, directly we have that the final term dominates for large $\|x-y\|_2$, and it follows that for $d > 1$, $\kummerkernel(x, y)$ decays like $\Gamma((d-1)/2)^{-1}\gamma^{-1/2}\|x-y\|_2^{-1}$.
\end{proof}
\kummermmdsample*
\begin{proof} \label{proof:kummermmdsample}
    The proof is a direct computation.
    By \cref{claim:kummermmd}, $\sigreg(\learnprob) = \mmd[\kummerkernel]^2(\learnprob, \mathcal{N}(0, \sigma^2))$, and we will show that $\mmdgauss[\kummerkernel]^2(\empprob{P}_n)$ is the (unbiased) U-statistic $\ummd[\kummerkernel]^2$ for $\mmd[\kummerkernel]^2(\learnprob, \mathcal{N}(0, \sigma^2))$.
    This U-statistic will be derived by integrating the contributions of $\mathcal{N}(0, 1)$ in closed form instead of sampling.
    By \cite{kernel-two-sample}, denoting $\gaussshorthand \defeq \mathcal{N}(0, \sigma^2)$, we have
    \begin{align*}
        \ummd[\kummerkernel]^2(\empprob{p}_n, \gaussshorthand)
        = \underbrace{\frac{1}{n(n-1)}\sum_{i=1}^n\sum_{j\neq i}\kummerkernel(\rv{x}_i, \rv{x}_j)}_{\mathrm{I}}
        + \underbrace{\iint\kummerkernel(y_1, y_2)\dd \gaussshorthand(y_1)\dd \gaussshorthand(y_2)}_{\mathrm{II} \defeq C}
        - \underbrace{\frac{2}{n}\sum_{i=1}^n\int \kummerkernel(\rv{x}_i, y)\dd \gaussshorthand(y)}_{\mathrm{III}}.
    \end{align*}

    Note that term $\mathrm{II}$ is totally independent of $\learnprob$, so we write it off as a constant $C$. The term $\mathrm{I}$ above is expressed solely in terms of pairwise kernel evaluations from samples from $\learnprob$, it will not be simplified further.
    To complete the proof, it remains to simplifty $\mathrm{III}$.
    Unpacking $\kummerkernel$ into an expectation over uniform random projections, we have
    \begin{align*}
        \mathrm{III}
        &= \frac{2}{n}\sum_{i=1}^n\int_{\mathbb{S}^{d-1}}\int_{\mathbb{R}}\exp(-\gamma(\direction^\top (\rv{x}_i - y))^2)\frac{1}{\sqrt{2\pi}\sigma}\exp(-\frac{1}{2\sigma^2}(\direction^\top y)^2)\,\dd y\dd\mathcal{U}(\direction)\\
        &\overset{(*)}{=} \frac{2}{n}\frac{1}{\sqrt{2\gamma\sigma^2}}\sum_{i=1}^n\int_{\mathbb{S}^{d-1}}\exp\left(-\frac{\gamma(\direction^\top \rv{x}_i)^2}{1 + 2\gamma\sigma^2}\right)\dd\mathcal{U}(\direction)\\
        &= \frac{2}{n\sqrt{1 + 2\gamma\sigma^2}}\sum_{i=1}^n\kummer\left(\frac{1}{2};\frac{d}{2};-\frac{\gamma}{1 + 2\gamma\sigma^2}\|\rv{x}_i\|_2\right).
    \end{align*}
    In the previous computation, step $(*)$ simply computes the expectation of $\exp(-\gamma \rv{z}^2)$, for $\rv{z} \defeq \direction^\top(\rv{x}_i - y)$, which is a Gaussian scalar random variable.
    Summing $\mathrm{I}$, this expression for $\mathrm{III}$, and $C$, we find that indeed $\mmdgauss[\kummerkernel]^2(\empprob{P}_n)$ is an unbiased estimator of $\sigreg(\learnprob)$ as a U-statistic of $\mmd[\kummerkernel]^2(\learnprob, \mathcal{N}(0,\sigma^2))$.

    For the final claim, it was shown in \cite[Theorem 6]{lejepa} that
    \begin{align*}
        \left\lvert\mathbb{E}\left[\sigregest(\empprob{P}_n) - \sigreg(\learnprob)\right]\right\rvert\leq O(1/n).
    \end{align*}

    Therefore, we immediately have
    \begin{align*}
        &\left\lvert\mathbb{E}\left[\mmdgauss[\kummerkernel]^2(\empprob{P}_n) - \sigregest(\empprob{P}_n)\right]\right\rvert\\
        &\quad=
        \left\lvert\mathbb{E}\left[\mmdgauss[\kummerkernel]^2(\empprob{P}_n) - \sigreg(\learnprob)\right] + \mathbb{E}\left[\sigreg(\learnprob) - \sigregest(\empprob{P}_n)\right]\right\rvert\\
        &\quad= \left\lvert\mathbb{E}\left[\sigreg(\learnprob) - \sigregest(\empprob{P}_n)\right]\right\rvert \leq O(1/n),\\
    \end{align*}
    which completes the proof.
\end{proof}

\thmcwmmdest*
\begin{proof}
    \label{proof:thmcwmmdest}
    The proof follows simply by substituting the result of \cref{thm:kummer-to-imq} into the result of \cref{claim:kummermmdsample} and applying the triangle inequality.
    Particularly, defining $\alpha \defeq \frac{4\gamma}{(2d-3)(1+2\gamma\sigma^2)}$, by \cref{thm:kummer-to-imq},
    \begin{align*}
        &\left\lvert\cwmmd(\empprob{p}_n) - \mmdgauss[\kummerkernel]^2(\empprob{p}_n)\right\rvert\\
        &\quad\leq \left\lvert\frac{1}{n(n-1)}\sum_{i=1}^n\sum_{j\neq i}\left[\kimq^{\nicefrac{4\gamma}{2d-3},\nicefrac{1}{2}}(\rv{x}_i, \rv{x}_j) - \kummer\left(\frac{1}{2};\frac{d}{2};-\gamma\|\rv{x}_i - \rv{y}_i\|_2^2\right)\right]\right\rvert\\
        &\qquad + \left\lvert\frac{2}{n\sqrt{1 + 2\gamma\sigma^2}}\sum_{i=1}^n \left[\frac{1}{\sqrt{1 + \alpha \|\rv{x}_i\|_2^2}} - \kummer\left(\frac{1}{2};\frac{d}{2};-\frac{\gamma}{1+2\gamma\sigma^2}\|\rv{x}_i\|_2^2\right)\right]\right\rvert\\
        &\quad \leq \frac{1}{n(n-1)}n(n-1)o_d(1) + \frac{2}{n\sqrt{1+ 2\gamma\sigma^2}}no_d(1)\\
        &\quad\leq o_d(1).
    \end{align*}
    Then, we have
    \begin{align*}
        &\lvert \mathbb{E}[\cwmmd(\empprob{P}_n) - \sigregest(\empprob{P}_n)]\rvert\\
        &\quad= \lvert \mathbb{E}[\cwmmd(\empprob{P}_n) -\mmdgauss[\kummerkernel]^2(\empprob{p}_n) + \mmdgauss[\kummerkernel]^2(\empprob{p}_n) - \sigregest(\empprob{P}_n)]\rvert\\
        &\quad\leq \lvert \mathbb{E}[\cwmmd(\empprob{P}_n) -\mmdgauss[\kummerkernel]^2(\empprob{p}_n)]\rvert + \lvert\mathbb{E}[\mmdgauss[\kummerkernel]^2(\empprob{p}_n) - \sigregest(\empprob{P}_n)]\rvert\\
        &\quad\leq o_d(1) + O(1/n)\\
        &\quad\leq O_d(1/n),
    \end{align*}
    where the penultimate step holds by \cref{claim:kummermmdsample}.
\end{proof}

\thmcwksdest*
\begin{proof}
    \label{proof:thmcwksdest}
    This result is a direct consequence of the expression for the sliced squared KSD given and proved in \cref{claim:slicedksd}.
    Indeed, $\cwksd$ is simply the U-estimator for the sliced square KSD, noting that
    \begin{align*}
        \kummer\left(\frac{1}{2};\frac{d}{2};-\gamma\|x - x\|_2^2\right)
        &= \kimq^{\nicefrac{4\gamma}{2d-3},\nicefrac{1}{2}}(x, x') + o_d(1)\\
        \kummer\left(\frac{3}{2};\frac{d}{2} + 1;-\gamma\|x - x\|_2^2\right)
        &= \kimq^{\nicefrac{2\gamma}{d},\nicefrac{3}{2}}(x, x') + o_d(1)
    \end{align*}
    as described earlier by \cite{cwae}.
\end{proof}

To complete the proof of \cref{claim:thmcwksdest}, we must establish an expression for the sliced squared KSD, which was used to derive the U-estimator above.

\begin{restatable}{lemma}{slicedksd}
\label{claim:slicedksd}
Let $\twosampledir{x}{x'} \defeq \|x - x'\|_2^{-1}(x - x')$.
Then the sliced KSD with kernel $\kgsn$ is given by
\begin{align*}
    &\sliceop\ksd[\kgsn]^2(\learnprob, \mathcal{N}(0, \sigma^2I_d))\\
    &\quad= \mathbb{E}_{\rv{x}, \rv{x}'\sim \learnprob}\left[\left(2\gamma + \frac{\rv{x}^\top \rv{x}' - \twosampledir{\rv{x}}{\rv{x}'}^\top \rv{x}\twosampledir{\rv{x}}{\rv{x}'}^\top \rv{x}'}{\sigma^4(d-1)}\right)\kummer\left(\frac{1}{2};\frac{d}{2};-\gamma \|\rv{x}-\rv{x}'\|_2^2\right)\right]\\
    &\qquad+ \mathbb{E}_{\rv{x}, \rv{x}'\sim \learnprob}\left[\frac{1}{d}\left(\frac{\twosampledir{\rv{x}}{\rv{x}'}^\top \rv{x}\twosampledir{\rv{x}}{\rv{x}'}^\top \rv{x}'}{\sigma^4} - \frac{\rv{x}^\top \rv{x}' - \twosampledir{\rv{x}}{\rv{x}'}^\top \rv{x}\twosampledir{\rv{x}}{\rv{x}'}^\top \rv{x}'}{\sigma^4(d-1)}\right)\kummer\left(\frac{3}{2};\frac{d}{2}+1;-\gamma \|\rv{x}-\rv{x}'\|_2^2\right)\right]\\
    &\qquad+ \mathbb{E}_{\rv{x}, \rv{x}'\sim \learnprob}\left[\frac{1}{d}\left(\frac{2\gamma}{\sigma^2} + 4\gamma^2\right)\|\rv{x} - \rv{x}'\|_2^2\kummer\left(\frac{3}{2};\frac{d}{2}+1;-\gamma \|\rv{x} - \rv{x}'\|_2^2\right)\right].
\end{align*}
\end{restatable}
\begin{proof}
    \label{proof:slicedksd}
    This proof will leverage several hypergeometric integral computations that are given in \cref{app:integrals}.
    Let $\gaussshorthand \sim \mathcal{N}(0,\sigma^2I_d)$. Note that for any $\direction\in\mathbb{S}^{d-1}$, we have $\mathrm{law}(\direction^\top \gaussshorthand) = \mathcal{N}(0, \sigma^2)$.
    Moreover, it is readily derived that the kernel $k_{\prob{q}}$ for the KSD as written in \eqref{eqn:ksd-expanded} for prior $\prob{q} = \Normal(0, \sigma^2)$ and base kernel $\kgsn$ is given by
    \begin{equation*}
        k_{\prob{q}} (u, v)
        = \left[\frac{uv}{\sigma^4} - \left(\frac{2\gamma}{\sigma^2} + 4\gamma^2\right)(u - v)^2 + 2\gamma\right]\exp(-\gamma(u - v)^2).
    \end{equation*}
    Recall that $\dif\mathcal{U}(\mathbb{S}^{d-1})(\direction) = \frac{\pi^{d/2}}{\Gamma(d/2)}\dif\direction \defeq C_d\dif\direction$.
    Now, the sliced squared KSD takes the following form,
    \begin{equation*}
    \begin{aligned}
        \sliceop\ksd[\kgsn]^2&(\learnprob, \mathcal{N}(0,\sigma^2I_d)) \\
        &= C_d\mathbb{E}_{\rv{x}, \rv{x}'\sim \learnprob}\int_{\mathbb{S}^{d-1}}\left(\frac{\direction^\top \rv{x} \direction^\top \rv{x}'}{\sigma^4} + 2\gamma\right)\exp(-\gamma(\direction^\top(\rv{x} - \rv{x}'))^2)\dif\direction\\
        &\quad - C_d\left(\frac{2\gamma}{\sigma^2} + 4\gamma^2\right)\mathbb{E}_{\rv{x}, \rv{x}'\sim P}\int_{\mathbb{S}^{d-1}}(\direction^\top(\rv{x} - \rv{x}'))^2\exp(-\gamma(\direction^\top(\rv{x} - \rv{x}'))^2)\dif\direction\\
        &\defeq C_d\mathbb{E}_{\rv{x}, \rv{x}'\sim \learnprob}\left[\mathrm{I}(\rv{x}, \rv{x}') + \left(\frac{2\gamma}{\sigma^2} + 4\gamma^2\right)\mathrm{II}(\rv{x}, \rv{x}')\right].
    \end{aligned}
    \end{equation*}
    Recall the functions $J_1, J_2$ defined in \cref{app:integrals}.
    Beginning with $\mathrm{I}$, we have
    \begin{align*}
        \mathrm{I}(x, x')
        &= \underbrace{\int_{\mathbb{S}^{d-1}}\frac{\direction^\top x\direction^\top x'}{\sigma^4}\exp(-\gamma(\direction^\top(x-x'))^2)\dif\direction}_{\mathrm{I}_1(x, x')} + 2\gamma J_1(\gamma\|x - x'\|_2^2).
    \end{align*}
    Using the closed form expression for $\mathrm{I}_1(x, x')$ derived in \cref{lem:slicedksd:terms:i:1}, we have
    \begin{align*}
        \mathrm{I}(x, x')
        &= \left(2\gamma + \frac{x^\top x' - e_1^\top x e_1^\top x'}{\sigma^4(d-1)}\right)J_1(\gamma\|x - x'\|_2^2)\\
        &\quad+ \left(\frac{e_1^\top x e_1^\top x'}{\sigma^4} - \frac{x^\top x' - e_1^\top xe_1^\top x'}{\sigma^4(d-1)}\right)J_2(\gamma\|x - x'\|_2^2),
    \end{align*}
    where $e_1 = \|x - x'\|_2^{-1}(x - x')$. Next, by \cref{lem:slicedksd:terms:ii}, we have
    \begin{align*}
        \mathrm{II}(x, x') = \|x - x'\|_2^2 J_2(\gamma\|x - x'\|^2).
    \end{align*}

    Putting these together, and substituting the closed forms for $J_1, J_2$ due to \cref{lem:integral:nasty:j1,lem:integral:nasty:j2}, we have
    \begin{talign*}
        &\sliceop\ksd^2(\learnprob, \mathcal{N}(0, \sigma^2))\\
        &\quad= C_d\mathbb{E}_{\rv{x}, \rv{x}'\sim \learnprob}\left[2\frac{\pi^{d/2}}{\Gamma(d/2)}\left(2\gamma + \frac{\rv{x}^\top \rv{x}' - e_1^\top \rv{x}e_1^\top \rv{x}'}{\sigma^4(d-1)}\right)\kummer\left(\frac{1}{2};\frac{d}{2};-\gamma \|\rv{x}-\rv{x}'\|_2^2\right)\right]\\
        &\qquad+ C_d\mathbb{E}_{\rv{x}, \rv{x}'\sim \learnprob}\left[\frac{\pi^{d/2}}{\Gamma(d/2 + 1)}\left(\frac{e_1^\top \rv{x}e_1^\top \rv{x}'}{\sigma^4} - \frac{\rv{x}^\top \rv{x}' - e_1^\top \rv{x}e_1^\top \rv{x}'}{\sigma^4(d-1)}\right)\kummer\left(\frac{3}{2};\frac{d}{2}+1;-\gamma \|\rv{x}-\rv{x}'\|_2^2\right)\right]\\
        &\qquad+ C_d\mathbb{E}_{\rv{x}, \rv{x}'\sim \learnprob}\left[\frac{\pi^{d/2}}{\Gamma(d/2 + 1)}\left(\frac{2\gamma}{\sigma^2} + 4\gamma^2\right)\|\rv{x} - \rv{x}'\|_2^2\kummer\left(\frac{3}{2};\frac{d}{2}+1;-\gamma \|\rv{x} - \rv{x}'\|_2^2\right)\right].
    \end{talign*}
    The claim follows by expanding $C_d$ and simplifying, noting that
    \begin{align*}
        \frac{\Gamma(d/2)}{\Gamma(d/2 + 1)} = \frac{\Gamma(d/2)}{\frac{d}{2}\Gamma(d/2)} = \frac{2}{d}.
    \end{align*}
\end{proof}
\section{Hypergeometric Integrals}\label{app:integrals}
In this section, we compute various integrals that will be useful throughout the text.
Henceforth, we fix $u, v\in\mathbb{R}^d$ and $\gamma>0$, and we define the following integrals:
\begin{equation*}\label{eq:integral:nasty:1}
J_1(c) := \int_{\mathbb{S}^{d-1}}\exp(-c\direction_1^2)\dd\direction,\quad c\in\mathbb{R}.
\end{equation*}
\begin{equation*}\label{eq:integral:nasty:2}
J_2(c) := \int_{\mathbb{S}^{d-1}}\direction_1^2\exp(-c\direction_1^2)\dd\direction,\quad c\in\mathbb{R}.
\end{equation*}

\begin{lemma}
    \label{lem:integral:nasty:j1}
    We have that
    \begin{align*}
        J_1(c) &= \frac{2\pi^{d/2}}{\Gamma(d/2)}\kummer\left(\frac{1}{2};\frac{d}{2};-c\right).
    \end{align*}
\end{lemma}
\begin{proof}
    This is the well-known closed form of the Gaussian integral on the sphere.
\end{proof}

\begin{lemma}
    \label{lem:integral:nasty:j2}
    We have that
    \begin{align*}
        J_2(c) &= \frac{\pi^{d/2}}{\Gamma(d/2 + 1)}\kummer\left(\frac{3}{2};\frac{d}{2}+1; -c\right).
    \end{align*}
\end{lemma}
\begin{proof}
    This follows directly from \cref{lem:slicedksd:terms:ii}.
\end{proof}

\begin{lemma}
    \label{lem:integral:nasty:orthogonal}
    Suppose $u, v$ are orthogonal to $e_1$, the first unit basis vector.
    Then
    \begin{align*}
        \int_{\mathbb{S}^{d-1}}(\direction^\top u)(\direction^\top v)\exp(-c\direction_1^2)\dd\direction
        = (u^\top v)\frac{J_1(c) - J_2(c)}{d-1}.
    \end{align*}
\end{lemma}
\begin{proof}
    Firstly, note that for $i, j\neq 1$ and $i\neq j$, we have
    \begin{align*}
        \int_{\mathbb{S}^{d-1}}\direction_i\direction_j\exp(-c\direction_1^2)\dd\direction = 0
    \end{align*}

    since the integrand is odd. Moreover, by symmetry, it is clear that
    \begin{align*}
        \int_{\mathbb{S}^{d-1}}\direction_i^2\exp(-c\direction_1^2)\dd\direction = \int_{\mathbb{S}^{d-1}}\direction_j^2\exp(-c\direction_1^2)\dd\direction
    \end{align*}
    for any $i, j\geq 2$. Since for any $\direction\in\mathbb{S}^{d-1}$ we have by definition $\sum_{i=1}^d\direction_i^2 = 1$, we have
    \begin{align*}
        J_1(c) &= \sum_{i=1}^d\int_{\mathbb{S}^{d-1}}\direction_i^2\exp(-c\direction_1^2)\dd\direction\\
        &= J_2(c) + (d-1)\int_{\mathbb{S}^{d-1}}\direction_i^2\exp(-c\direction_1^2)\dd\direction && i\geq 2\\
        \int_{\mathbb{S}^{d-1}}\direction_i^2\exp(-c\direction_1^2)\dd\direction &= \frac{J_1(c) - J_2(c)}{d-1}.
    \end{align*}
    We can now relate this result to the integral we wish to compute. Particularly, we have
    \begin{align*}
        \int_{\mathbb{S}^{d-1}}(\direction^\top u)(\direction^\top v)\exp(-c\direction_1^2)\dd\direction
        &= \sum_{i=2}^d\sum_{j=2}^d\int_{\mathbb{S}^{d-1}}u_iv_j\direction_i\direction_j\exp(-c\direction_1^2)\dd\direction\\
        &= \sum_{i=2}^du_iv_i\int_{\mathbb{S}^{d-1}}\direction_i^2\exp(-c\direction_1^2)\dd\direction\\
        &= u^\top v \frac{J_1(c) - J_2(c)}{d-1}.
    \end{align*}
\end{proof}

\begin{lemma}
    \label{lem:slicedksd:terms:i:1}
    Let $x, x'\in\mathbb{R}^d$. Then define $\mathrm{I}_1(x, x')$ via
    \begin{align*}
        \mathrm{I}_1(x, x') &= \int_{\mathbb{S}^{d-1}}(\direction^\top x)(\direction^\top x')\exp(-\gamma(\direction^\top(x - x'))^2)\dif\direction.
    \end{align*}
    It holds that
    \begin{align*}
        \mathrm{I}_1(x, x') = (x^\top x' - (e_1^\top x)(e_1^\top x'))\frac{J_1(r\gamma^2) - J_2(r\gamma^2)}{d-1} + (e_1^\top x)(e_1^\top x')J_2(r\gamma^2).
    \end{align*}
\end{lemma}
\begin{proof}
    Let $\delta = x - x'$, and consider a coordinate system in which $\delta$ is parallel to the first basis vector denoted $e_1$. Denoting by $x_\bot$ the component of $x$ orthogonal to $e_1$, we have
    \begin{align*}
        \direction^\top x &= \direction_1(e_1^\top x) + \direction^\top x_\bot\\
        \direction^\top x' &= \direction_1(e_1^\top x') + \direction^\top x'_\bot.
    \end{align*}
    Letting $r = \|x - x'\|_2$, we have
    \begin{align*}
        \mathrm{I}_1(x, x')
        &= \underbrace{\int_{\mathbb{S}^{d-1}}\direction_1^2(e_1^\top x)(e_1^\top x')\exp(-\gamma r^2\direction_1^2)\dif\direction}_{f_1(x, x')}\\
        &\quad+ \int_{\mathbb{S}^{d-1}}\direction_1(e_1^\top x)(\direction^\top x'_\bot)\exp(-\gamma r^2\direction_1^2)\dif\direction\\
        &\quad+ \int_{\mathbb{S}^{d-1}}\direction_1(e_1^\top x')(\direction^\top x_\bot)\exp(-\gamma r^2\direction_1^2)\dif\direction\\
        &\quad+ \underbrace{\int_{\mathbb{S}^{d-1}}(\direction^\top x_\bot)(\direction^\top x'_\bot)\exp(-\gamma r^2\direction_1^2)\dif\direction}_{f_2(x, x')}.
    \end{align*}
    Note that the integrands in the middle two terms are odd, so those integrals vanish.
    Moreover, we have
    \begin{align*}
        f_1(x, x') &\equiv (e_1^\top x)(e_1^\top x')J_2(\gamma r^2).
    \end{align*}

    Furthermore, by \cref{lem:integral:nasty:orthogonal}, we have
    \begin{align*}
        f_2(x, x') = x_\bot^\top x'_\bot \frac{J_1(\gamma r^2) - J_2(\gamma r^2)}{d-1}.
    \end{align*}
    Altogether, we have shown that
    \begin{align*}
        \mathrm{I}_1(x, x')
        &= (x^\top x' - (e_1^\top x)(e_1^\top x'))\frac{J_1(r\gamma^2) - J_2(r\gamma^2)}{d-1} + (e_1^\top x)(e_1^\top x')J_2(r\gamma^2).
    \end{align*}
\end{proof}

\begin{lemma}
    \label{lem:slicedksd:terms:ii}
    Let $x, x'\in\mathbb{R}^d$. Then it holds that
    \begin{align*}
        \mathrm{II}(x, x')
        &:= \int_{\mathbb{S}^{d-1}}(\direction^\top(x - x'))^2\exp(-\gamma(\direction^\top(x - x'))^2)\dif\direction\\
        &= \|x - x'\|_2^2\frac{\pi^{d/2}}{\Gamma(d/2 + 1)}\kummer\left(\frac{3}{2}; \frac{d}{2} + 1; -\gamma\|x - x'\|_2^2\right).
    \end{align*}
\end{lemma}
\begin{proof}
    Let $\delta = x - x'$. Immediately, overloading notation, we have $\mathrm{II}(x, x')\equiv \mathrm{II}(\delta)$, where
    \begin{align*}
        \mathrm{II}(\delta)
        &:= \int_{\mathbb{S}^{d-1}}(\direction^\top\delta)^2\exp(-\gamma(\direction^\top\delta)^2)\dif\direction.
    \end{align*}
    Note that $\mathrm{II}$ is rotationally symmetric. So, without loss of generality, we can take $\delta = re_1$, where $r = \|\delta\|_2$ and $e_1$ is the first standard unit vector in $\mathbb{R}^d$. Then, we have
    \begin{align*}
        \mathrm{II}(\delta)
        &= \int_{\mathbb{S}^{d-1}}(r\direction_1)^2\exp(-\gamma (r\direction_1)^2)\dif\direction\\
        &= r^2\int_{\mathbb{S}^{d-1}}\direction_1^2\exp(-\gamma (r\direction_1)^2)\dif\direction.
    \end{align*}
    Now we've expressed $\mathrm{II}$ as an integral of a function depending on only one coordinate over the unit sphere. Switching to polar coordinates, defining $\direction_1 = \cos\phi$, we have
    \begin{align*}
        \mathrm{II}(\delta)
        &= r^2\mathrm{Area}(\mathbb{S}^{d-2})\int_0^\pi\cos^2\phi\exp(-\gamma(r\cos\phi)^2)\sin^{d-2}(\phi)\dif\phi\\
        &= r^2\frac{2\pi^{(d-1)/2}}{\Gamma((d-1)/2)}\int_0^\pi\cos^2\phi\exp(-\gamma(r\cos\phi)^2)\sin^{d-2}(\phi)\dif\phi.
    \end{align*}
    Now, writing $u := \direction_1$, we have $\sin^{d-2}(\phi)\dif\phi = -(1 - u^2)^{(d-3)/2}\dif u$, so that
    \begin{align*}
        \mathrm{II}(\delta)
        &= r^2\frac{2\pi^{(d-1)/2}}{\Gamma((d-1)/2)}\int_{-1}^{1}u^2(1-u^2)^{(d-3)/2}\exp(-\gamma r^2 u^2)\dif u\\
        &= 2r^2\frac{2\pi^{(d-1)/2}}{\Gamma((d-1)/2)}\int_{0}^{1}u^2(1-u^2)^{(d-3)/2}\exp(-\gamma r^2 u^2)\dif u.
    \end{align*}
    We will now express this in terms of $\kummer$. Recall that
    \begin{align*}
        \kummer(a; b; z) &= \frac{\Gamma(b)}{\Gamma(a)\Gamma(b - a)}\int_0^1 t^{a-1}(1 - t)^{b - a - 1}e^{zt}\dif t.
    \end{align*}
    To relate $\mathrm{II}(\delta)$ to $\kummer$, we write $t := u^2$, so that $\dif u = \frac{1}{2}t^{-1/2}\dif t$, yielding
    \begin{align*}
        \mathrm{II}(\delta)
        &= r^2\frac{2\pi^{(d-1)/2}}{\Gamma((d-1)/2)}\int_{0}^{1}t^{1/2}(1-t)^{(d-3)/2}\exp(-\gamma r^2 t)\dif t.
    \end{align*}
    Now let $a = 3/2$, $b = d/2 + 1$, and $z = -\gamma r^2$, so that
    \begin{align*}
        \mathrm{II}(\delta)
        &= r^2\frac{2\pi^{(d-1)/2}}{\Gamma((d-1)/2)}\int_{0}^{1}t^{a - 1}(1-t)^{b - a - 1}\exp(z t)\dif t\\
        &= r^2\frac{2\pi^{(d-1)/2}}{\Gamma((d-1)/2)}\left[\left(\frac{\Gamma(d/2 + 1)}{\Gamma(3/2)\Gamma((d-3)/2)}\right)^{-1}\kummer\left(\frac{3}{2}; \frac{d}{2} + 1; z\right)\right]\\
        &= \|x - x'\|_2^2\frac{\pi^{d/2}}{\Gamma(d/2 + 1)}\kummer\left(\frac{3}{2}; \frac{d}{2} + 1; -\gamma\|x - x'\|_2^2\right).
    \end{align*}
\end{proof}

\end{document}